\DeclareUnicodeCharacter{2061}{}
\documentclass[journal]{IEEEtran}

\makeatletter
\def\ps@headings{%
	\def\@oddhead{\mbox{}\scriptsize\rightmark \hfil \thepage}%
	\def\@evenhead{\scriptsize\thepage \hfil \leftmark\mbox{}}%
	\def\@oddfoot{}%
	\def\@evenfoot{}}
\makeatother 

\usepackage{graphicx}
\usepackage{amsmath, amsfonts,epsfig, multirow, floatflt}
\usepackage{amssymb}
\usepackage{subfigure}
\usepackage{cite}
\usepackage{algorithm,algorithmic}
\usepackage{hyperref}
\usepackage{enumitem}
\usepackage{diagbox}
\usepackage{mathrsfs}
\usepackage{url}
\usepackage{color}
\usepackage{xcolor}

\graphicspath{{figure/}}

\begin{document}
	
\title{Dynamic Energy Dispatch Based on Deep Reinforcement Learning in IoT-Driven Smart Isolated Microgrids }

\author{Lei~Lei {\it Senior Member, IEEE}, Yue~Tan, Glenn~Dahlenburg, Wei~Xiang {\it Senior Member, IEEE}, Kan~Zheng {\it Senior Member, IEEE}

}

\maketitle

\begin{abstract}
Microgrids (MGs) are small, local power grids that can operate independently from the larger utility grid. Combined with the Internet of Things (IoT), a smart MG can leverage the sensory data and machine learning techniques for intelligent energy management. This paper focuses on deep reinforcement learning (DRL)-based energy dispatch for IoT-driven smart isolated MGs with diesel generators (DGs), photovoltaic (PV) panels, and a battery. A finite-horizon Partial Observable Markov Decision Process (POMDP) model is formulated and solved by learning from historical data to capture the uncertainty in future electricity consumption and renewable power generation. In order to deal with the instability problem of DRL algorithms and unique characteristics of finite-horizon models, two novel DRL algorithms, namely, finite-horizon deep deterministic policy gradient (FH-DDPG) and finite-horizon recurrent deterministic policy gradient (FH-RDPG), are proposed to derive energy dispatch policies with and without fully observable state information. A case study using real isolated MG data is performed, where the performance of the proposed algorithms are compared with the other baseline DRL and non-DRL algorithms. Moreover, the impact of uncertainties on MG performance is decoupled into two levels and evaluated respectively.  	  
\end{abstract}

	
	\section{Introduction}
	Microgrids (MGs) are small-scale, self-supporting power networks driven by on-site generation sources. They can potentially integrate renewable energy sources (RESs) such as solar and wind, as well as energy storage elements such as electrochemical battery, and often use a base load generation source such as reciprocating diesel engines of heavy fuel oil. An MG can either connect or disconnect from the external main grid to operate in grid-connected or isolated-mode. MGs improve grid reliability and supply sustainable and quality electric power. However, their planning and operations are faced with challenges from uncertainty because of the difficulty in accurate prediction of future electricity consumption and renewable power generation. As RESs reach high levels of grid penetration, intelligent energy management that can handle variability and uncertainty becomes essential for an MG to provide reliable power supply efficiently.\par
	
	In recent years, Internet of Things (IoT) play a major role in turning the MG systems into smart MGs. The rapidly growing applications of IoT devices and technologies contribute to the achievement of the MGs' goals to reduce non-renewable energy cost, satisfy the consumers' demand, and maintain the balance between power demand and generation \cite{li2018implemented,li2018real}. Intelligent energy management needs data from throughout the MGs, while IoT makes it easier to collect and analyze data from traditional grid resources, RESs and other electrical assets. With this data, IoT-driven smart MGs can perform intelligent energy management by leveraging machine learning techniques to handle variability and uncertainty \cite{bedi2018review,minoli2017iot}. IoT-driven smart grid is a typical application scenario of autonomous IoT systems, where IoT devices with sensors and actuators can interact with the environment to acquire data and exert control actions \cite{Lei2019}. \par

	
    In this paper, we focus on an IoT-driven smart isolated MG system including diesel generators (DGs), photovoltaic (PV) and a battery with control capabilities, based on connected sensors, deep reinforcement learning (DRL) and other IoT technologies. Isolated MG systems can be found in many remote areas all over the world, which are inefficient to connect to the main grid. As the MGs in remote locations usually lack advanced control hardware, we do not consider controllable load for demand side management. However, a load bank is used at the power station to protect the DGs, so that excessive DG power can be dissipated to avoid curtailing the generation. We consider the DGs as the primary source of power, while the PV acts as a clean complementary power source. Our main objective is to manage the energy dispatch of the DGs so that their operational cost can be minimized by fully exploiting the PV power, while satisfying the critical requirement to maintain balance between power generation and consumption. The main challenge stems from the unpredictability in future PV generation and load demand, making it difficult to maintain balance between power generation and consumption without inefficient operations of the DGs. \par
    
    The battery can help overcome the above challenge to some extent, as it can charge or discharge to compensate for short-term unbalance. Moreover, the battery can act as an additional power source to minimize operational cost of DGs. However, the battery also brings new challenges in energy management. As the maximum amount of energy that can be charged or discharged at a certain point of time is limited by the energy storage capability and the current state-of-charge (SoC) of the battery, while the current SoC is in turn determined by the previous charging/discharging behavior, energy management becomes a sequential decision problem for a dynamic system where earlier decisions influence future available choices. Therefore, dynamic optimization techniques are required to optimize the overall performance over a finite time horizon instead of only the instantaneous performance at a single point of time. \par

    \begin{table*} [!h]
    	\centering
    	\renewcommand{\arraystretch}{1.5}
    	\caption{Approaches for Energy Management in Microgrids and Some Example References.}
    	\label{table_summary}
    	{    
    		\begin{tabular}{|p{4.3cm}|p{1.5cm}|p{8.5cm}|}
    			\hline
    			Approach & References & Description \\
    			\hline
    			Stochastic Programming (SP) &\cite{Su2014,Wang2015,Farzin2017} & find a policy that maximizes the expected value for a range of possible scenarios\\
    			\hline
    			Robust Optimization (RO) & \cite{Zhang2013,Giraldo2019} & solves a deterministic problem over an uncertainty set to derive a solution that is feasible for any realization within the uncertainty set\\
    			\hline
    			Model Predictive Control (MPC) / Receding Horizon Control (RHC) & \cite{Olivares2014,Zhu2014,Sachs2016,Lara2019} &  iteratively solve optimal control problems for the prediction horizon  \\
    			\hline
    			Approximate Dynamic Programming (ADP) & \cite{Zeng2018,Shuai2018,Shuai2019} & solve MDP problems and overcome curse-of-dimensionality by function approximation \\
    			\hline
    			Lyapunov Optimization (LO) & \cite{Shi2017,Hu2018} & stabilize queueing networks
    			while additionally optimize some performance metrics \\
    			\hline
    			Deep Reinforcement Learning (DRL) & \cite{Francois-Lavet2016,Ji2019,Venayagamoorthy2016,Foruzan2018,yu2019deep,munir2019edge} & combination of RL and DL to solve the MDP problems without requiring the stochastic properties of the underlying MDP model \\
    			\hline
    		\end{tabular}}
    	
    \end{table*}

	\subsection{Related Work}
	\subsubsection{Energy Management Approaches in MG}
    Several approaches to dealing with uncertainties have been adopted for energy management in MGs by recent works. Table \ref{table_summary} lists some typical approaches, with a brief description for each approach and some example references adopting the approach. All the approaches solve sequential stochastic optimization problems with theories from competing fields such as Markov decision processes (MDP), stochastic programming, and optimal control. We refer to \cite{Powell2016,Powell2016a} for a clear explanation of the relationship between different approaches. \par
    
    \textit{Stochastic Programming} (\textit{SP}) is a field that evolved out of deterministic mathematical programming. In SP, the objective functions are usually in the form of expected values for a range of possible scenarios, where the probabilities for these scenarios to occur are considered to be available. In \cite{Su2014,Wang2015}, the expected operational cost of the MG is minimized while accommodating the intermittent nature of renewable energy resources. In \cite{Farzin2017}, a multi-objective stochastic optimization problem is formulated for optimal scheduling of batteries in a MG. In \textit{Robust Optimization} (\textit{RO}), the goal is to make a decision that is optimal for the worst-case objective function. In \cite{Zhang2013}, a robust formulation accounting for the worst-case amount of harvested RESs is developed for distributed economic dispatch in a MG. In \cite{Giraldo2019}, a robust convex optimization model is proposed for the energy management of MGs, where global robustness is attained by a single setting parameter.\par
    
    In \textit{Model Predictive Control} (\textit{MPC})/Receding Horizon Control (\textit{RHC}), the optimal control problem for a dynamic system is considered. At each time step, a sequence of optimal control actions is computed by solving an open-loop deterministic optimization problem for the prediction horizon, but only the first value of the computed control sequence is implemented. This process is iteratively repeated, and a prediction model of the dynamic system behavior is required. In \cite{Olivares2014}, the \textit{MPC} technique is applied to determine the unit commitment and optimal power flow problems in an isolated MG, where the stochastic variables are assumed to be perfectly predicted. As the performance of the \textit{MPC} based approaches heavily depends on the accuracy of the prediction model, a stochastic \textit{MPC} method is adopted in \cite{Zhu2014} to optimally dispatch energy storage and generation, where the uncertain variables are treated as random processes. In \cite{Sachs2016}, an \textit{MPC} strategy for isolated MGs in rural areas is developed, where the first stage derives the optimal power dispatch based on real-time predication of future power profiles, while the second stage adjusts the DG power to improve the robustness of the control strategy toward prediction errors. In \cite{Lara2019}, \textit{RO} and \textit{RHC} are combined to manage forecasting errors in energy management of isolated MGs. \par
    
    \textit{Dynamic Programming} (\textit{DP}) normally considers stochastic optimal control problems, where the system states evolve over time in a stochastic manner. The stochastic evolution of the state-action pair over time forms a MDP. \textit{DP} is a model-based control technique similar to MPC, which requires the transition probabilities of the MDP model to be available. One important challenge in DP is the curse-of-dimensionality problem, where the state space of the MDP model is too large to derive solutions within a reasonable period of time. Therefore, \textit{Approximate Dynamic Programming} (\textit{ADP}) is widely adopted as an efficient method to overcome this challenge by approximating the value and policy functions. In \cite{Zeng2018}, an \textit{ADP} based approach is developed to derive real-time energy dispatch of an MG, where the value functions are learned through neural networks. A near optimal policy is obtained through the approximate policy iteration algorithm. In \cite{Shuai2018,Shuai2019}, the optimal operation of MG is formulated as a stochastic mixed-integer nonlinear programming (MINLP) problem, and \textit{ADP} is applied to decompose the original multi-time-period MINLP problem into single-time period nonlinear programming problems. \par
    
    In \textit{Lyapunov Optimization} (\textit{LO}), the optimal control of a dynamic system is obtained through the Lyapunov function. A typical goal is to stabilize queueing networks while additionally optimize some performance metrics. In \cite{Shi2017}, the online energy management for real-time operation of MGs is modeled as a stochastic optimal power flow problem based on \textit{LO}. In \cite{Hu2018}, a two-stage optimization approach is proposed, where the first stage of optimization determines hourly unit commitment via a day-ahead scheduling, and the second stage performs economic dispatch and energy trading via a real-time scheduling based on \textit{LO}. \par


%
    \subsubsection{Deep Reinforcement Learning}
    In the above papers, the stochastic variables and processes are either assumed to be available by prediction models or represented by their corresponding expected values. On the other hand, Reinforcement Learning (RL) provides model-free methods to solve the optimal control problems of dynamic systems without requiring the stochastic properties of the underlying MDP model \cite{Zheng2016,ODonoghue2017,Heess2015a}. When combined with Deep Learning (DL), the more powerful DRL methods can deal with the curse-of-dimensionality problem by approximating the value functions as well as policy functions using deep neural networks \cite{Lei2019}. \par 
    
    RL/DRL algorithms can be broadly classified into value-based method, such as DQN \cite{Mnih2015} and Double DQN \cite{VanHasselt2016}; Monte Carlo policy gradient method, such as REINFORCE \cite{williams1992}; and actor-critic method, such as Stochastic Value Gradients (SVG)\cite{Heess2015a}, Deep Deterministic Policy Gradient (DDPG) \cite{lillicrap2015}, asynchronous Advantage Actor-Critic (A3C) \cite{mnih2016}, Trust Region Policy Optimization (TRPO) \cite{schulman2015}, Recurrent Deterministic Policy Gradients (RDPG) \cite{heess2015}. Actor-critic method combines the advantages of both value-based and Monte Carlo policy gradient methods. Compared with the Monte Carlo policy gradient method, it requires a far less number of samples to learn from and less computational resources to select an action, especially when the action space is continuous. Compared with the value-based method, it can learn stochastic policies and solve RL problems with continuous actions. However, actor-critic method may be unstable due to the recursive use of value estimates. \par
    
    Recent years have seen emerging applications of RL/DRL to provide energy management solutions for MGs \cite{Francois-Lavet2016,Ji2019,Venayagamoorthy2016,Foruzan2018,yu2019deep,munir2019edge}. In \cite{Francois-Lavet2016}, a value-based DRL algorithm is proposed to optimally activating the energy storage devices, where three discrete actions are considered, i.e., discharge at full rate, keep it idle, charge at full rate. Value-based DRL algorithms are relatively simple, but it cannot deal with continuous actions that are common in energy management problems. In \cite{Ji2019}, a DQN-based approach is used for real-time scheduling of an MG considering the uncertainty of the load demand, renewable energy, and electricity price. In order to apply the value-based DQN algorithm, the continuous action space is discretized, which results in performance loss. In \cite{Venayagamoorthy2016}, an evolutionary adaptive {color{red}DP} and RL framework is introduced to develop an intelligent dynamic energy management system for a smart MG. The proposed solution has a similar architecture to that of actor-critic DRL algorithms. It is considered in \cite{Venayagamoorthy2016} that the amount of renewable energy generation and load demand during each time step is available at the beginning of that time step, which is not the practical case. In \cite{Foruzan2018}, a multi-agent-based RL model is used to study distributed energy management in an MG. Considering the existence of model uncertainty and parameter constraints, a DDPG-based DRL algorithm is proposed in \cite{yu2019deep} to solve the energy management problem for smart home. In \cite{munir2019edge}, the authors focus on MG-enabled multi-access edge computing networks and formulate the energy supply plan problem into MDP, which is solved by the proposed DRL-based approach.\par 
    
    None of the above research works focus on the energy dispatch problem in an IoT-driven smart isolated MG. Moreover, DRL algorithms are known to be unstable and hard to converge, and there are many challenges when applied to solve real-world problems \cite{rlblogpost}. Furthermore, in the energy management problems for MGs, the system performance is usually optimized over a finite period of time, e.g., one day or one month. However, many existing DRL algorithms are developed without considering the difference between the finite-horizon MDP problems and the infinite-horizon MDP problems. Therefore, it is important to develop stable and efficient DRL algorithms for energy management in isolated MGs to optimize the system performance over a finite period of time.  \par

   	\subsection{Contributions} 
   	In this paper, we formulate a finite-horizon Partial Observable Markov Decision Process (POMDP) model for the energy dispatch of DGs in smart isolated MGs. For comparison purpose, we also formulate a finite-horizon MDP model to study the impact of uncertainty on system performance. We focus on intra-day operation of the MG where the duration of one day is divided into multiple time steps. The energy dispatch decision for DGs is determined at the beginning a time step. The objective is to optimize the sum of performance over all the time steps within the time horizon, i.e., one day. Performance considered in the optimization problem includes: (1) supply the power requirements of customer loads at all times; (2) eliminate reverse power flow where power generation is larger than power requirements of loads; (3) minimize power generation cost of DGs and maximize the utilization of renewable energy resource. Note that it is straightforward to adapt the proposed finite-horizon models to study intra-month and intra-year operations.  \par
   	
   	The main contributions of this paper lie in the following aspects:
   	\begin{itemize}
   		\item\emph{\textbf{DRL models that facilitate analyzing the impact of uncertainty}}: In this paper, the proposed finite-horizon MDP and POMDP models enable one to analyze and address the impact of uncertainty on smart isolated MGs. Specifically, we divide the uncertainty into two time granularities: (1) uncertainty due to data variation between two consecutive time steps; (2) uncertainty due to data variation between two consecutive days. The ``partial observable" and ``finite-horizon" formulation of the DRL models help us to capture the two levels of uncertainty, respectively. 
   		
   		\item \textbf{\emph{DRL algorithms that address the instability problem and finite-horizon setting}}: In this paper, we design a DRL algorithm namely finite-horizon deep deterministic policy gradient (FH-DDPG) for finite-horizon MDP model based on a well-known DRL algorithm, i.e., DDPG, which was demonstrated to be an effective method for training deep neural network policies \cite{Duan2016}, and has been adopted to solve problems in various areas \cite{yu2019deep,Qiu2019}. The instability problem of DRL algorithms and unique characteristics of finite-horizon setting are addressed using two key ideas - backward induction and time-dependent actors. While the idea of time-dependent actors has been studied for actor-critic RL in \cite{Grondman2013}, we apply it to two well-known DRL algorithms, i.e., DDPG and RDPG, and demonstrate its efficiency in dealing with finite-horzion problems. We demonstrate that the proposed FH-DDPG algorithm outperforms the baseline DDPG algorithm by large margins and is much more stable for solving the energy management problem.
   		
   		\item \textbf{\emph{DRL algorithms that address the partial observable problem}}: In order to address the partial observable problem, we develop a second DRL algorithm namely finite-horizon recurrent deterministic policy gradient (FH-RDPG) for finite-horizon POMDP model. Similar to RDPG, which is a well-known DRL algorithm specifically designed to work in the partial observable domain, FH-RDPG also incorporates Long Short Term Memory (LSTM) layers, which is a widely adopted recurrent neural network (RNN) architecture capable of capturing long term time dependencies. However, the overall workflow of the FH-RDPG algorithm is similar to that of the FH-DDPG algorithm, both of which are based on backward induction. We specifically define the history in the POMDP model for the MG environment to facilitate the implementation of the FH-RDPG algorithm. We demonstrate that the proposed FH-RDPG algorithm outperforms the baseline RDPG algorithm in an POMDP environment, where the PV generation and load for the current time step is not available. 
   	\end{itemize}

   	  
    The remainder of the paper is organized as follows. The system model is introduced in Section II. Section III formulates the MDP and POMDP models, which are solved by the two proposed DRL algorithms introduced in Section IV. In Section V, the performance of the proposed algorithms are compared with those of other baseline algorithms by simulation, where the results are analyzed and discussed. Section VI concludes the paper. 
	
	\section{System Model}
\begin{figure}[!htb]
		\centering
		\includegraphics[width=0.48\textwidth]{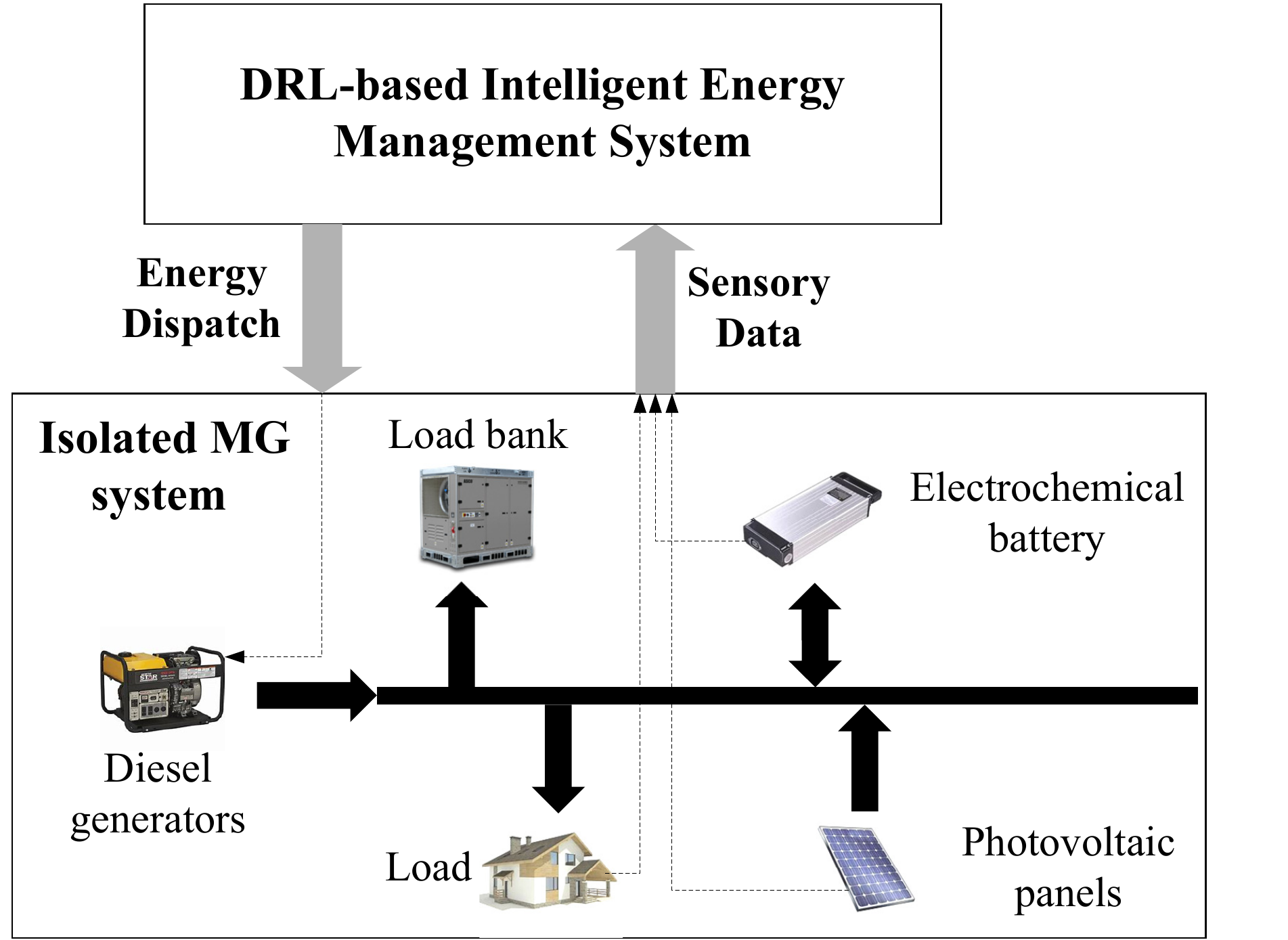}
		\caption{Schematic diagram of an IoT-driven smart isolated MG system.}
		\label{system_model}
	\end{figure}

We consider an IoT-driven smart isolated MG system, which consists of DGs, several PV panels, an electrochemical battery for energy storage, loads, and a load bank as shown in Fig. \ref{system_model}. Different from the traditional MG system, the IoT-driven smart MG comprises a DRL-based intelligent energy management system. The sensors deployed in the isolated MG collect sensory data to reflect the status of the MG system, which includes the output power generated by the PV panels, the load power demand, as well as the SoC of the battery. Based on the above real-time sensory data, the DRL-based intelligent energy management system makes dynamic energy dispatch decisions to drive the DGs in the isolated MG.  \par
 
The intra-day operation of the MG is divided into $T$ time steps, indexed by $\{1,\cdots,T\}$. The interval for each time step is $\Delta t$.

    \subsection{Diesel Generator Model}
    We consider there are $D$ DGs $(DG_{1},\cdots,DG_{D})$. Let $P_{t}^{DG_{d}}$ denote the output power of the $d$-th DG $DG_{d}$, $\forall d\in\{1,\cdots,D\}$, at time step $t$. The operational constraints are given by 
    \begin{equation}
    \label{eq1}
    P_{\mathrm{min}}^{DG_{d}}\leq P_{t}^{DG_{d}}\leq P_{\mathrm{max}}^{DG_{d}}, \forall d\in\{1,\cdots,D\},
    \end{equation}
    \noindent where $P_{\mathrm{min}}^{DG_{d}}$ and $P_{\mathrm{max}}^{DG_{d}}$ are the minimum and maximum output powers of $DG_{d}$, respectively. \par

    Let $c_{t}^{DG_{d}}$ be the generation cost of $DG_{d}$, which can be derived by the conventional quadratic cost function \cite{Zeng2018}
    \begin{equation}
    \label{eq2}
    c_{t}^{DG_{d}}=[a_{d}(P_{t}^{DG_{d}})^{2}+b_{d}P_{t}^{DG_{d}}+c_{d}]\Delta t, \forall d\in\{1,\cdots,D\},
    \end{equation} 
    \noindent where $a_{d}$, $b_{d}$, and $c_{d}$ are positive coefficients for $DG_{d}$.\par

    \subsection{Battery Model}
    Let $E_{t}$ denote the SoC of the battery at the beginning of time step $t$. Let $P_{t}^{\mathrm{E}}$ denote the charging or discharging power of the battery, and $u_{t}$ indicate charging status, which is $1$ if battery is charging and 0 otherwise.\par

    According to a simplified book-keeping model for the SoC of battery \cite{Bergveld2001,Olivares2014,Lara2019}, $E_{t+1}$ at time step $t+1$ can be derived based on the SoC $E_{t}$ at time step $t$ as
    \begin{equation}
    \label{eq3}
    E_{t+1}=E_{t}+\eta_{\mathrm{ch}} u_{t} P_{t}^{\mathrm{E}} \Delta t-(1-u_{t})P_{t}^{\mathrm{E}}\Delta t/\eta_{\mathrm{dis}},
    \end{equation}
    \noindent where $\eta_{\mathrm{ch}}$ and $\eta_{\mathrm{dis}}$ are the charging and discharging efficiencies of the battery \cite{Zeng2018}.
    
    In order to determine $P_{t}^{\mathrm{E}}$ and $u_{t}$, we define the following variable
    \begin{equation}
    \label{eq4}
	\delta_{t}=\sum_{d=1}^{D}P_{t}^{\mathrm{DG}_{d}}+P_{t}^{\mathrm{PV}}-P_{t}^{\mathrm{L}},
	\end{equation}
	\noindent where $P_{t}^{\mathrm{PV}}$ and $P_{t}^{\mathrm{L}}$ are the aggregated PV output power and load power demand at time step $t$, respectively.
	
	Therefore, we can set
    \begin{equation}
    \label{eq5}
	u_{t}=\left\{
	\begin{array}{ll}
	1, & \mathrm{if} \ \delta_{t}\geq 0\\
	0, & \mathrm{if} \ \delta_{t}<0 \\
	\end{array}\right. ,
	\end{equation}
	
	\noindent and
    \begin{equation}
    \label{eq6}
	P_{t}^{\mathrm{E}}=\left\{
	\begin{array}{ll}
	\min(\delta_{t},P_{\mathrm{ch\_lim}}^{\mathrm{E}}), & \mathrm{if} \ \delta_{t}\geq 0\\
	\min(-\delta_{t},P_{\mathrm{dis\_lim}}^{\mathrm{E}}), & \mathrm{if} \ \delta_{t}<0 \\
	\end{array}\right. ,
	\end{equation}			
	
	\noindent where
   \begin{equation}
   \label{eq7}	
	P_{\mathrm{ch\_lim}}^{\mathrm{E}}=\min⁡(P_{\mathrm{max}}^{\mathrm{E}},(E_{\mathrm{max}}-E_{t})/(\eta_{\mathrm{ch}}\Delta t)),
	\end{equation}	
	\noindent and
	\begin{equation}
	\label{eq8}	
	 P_{\mathrm{dis\_lim}}^{\mathrm{E}}=\min⁡(P_{\mathrm{max}}^{\mathrm{E}},\eta_{\mathrm{dis}}(E_{t}-E_{\mathrm{min}})/\Delta t),
	\end{equation}
	\noindent are the battery charging and discharging power limit. $E_{\mathrm{max}}$ and $E_{\mathrm{min}}$ are the maximum and minimum energy level of the battery, and $P_{\mathrm{max}}^{\mathrm{E}}$ is the maximum charging or discharging power. \par
	
	If $\delta_{t}> 0$, the generation is larger than demand, the excessive power will be charged to the battery. However, if $\delta_{t}>P_{\mathrm{ch\_lim}}^{\mathrm{E}}$, the excessive power beyond the charging capability of the battery will go to the load bank, which will be lost or generation will be curtailed. If $\delta_{t}<0$, the generation is smaller than demand, the battery needs to be discharged to supply the load. However, if $-\delta_{t}>P_{\mathrm{dis\_lim}}^{\mathrm{E}}$, even the battery cannot provide enough power, which will see part of the load unserved. \par

	\section{MDP and POMDP Models}
	In this section, we shall formulate a finite-horizon POMDP problem to minimize the power generation cost as well as the power demand and generation unbalance, where the fluctuating loads and stochastic generation of PV panels are taken into account. In order to analyze the impact of uncertainty on system performance, we also formulate a corresponding finite-horizon MDP problem for comparison purpose.  \par
	
    \subsection{State and Observation}
    We define the system state at time step $t$, $\forall t\in\{1,2,\cdots,T\}$, as $s_{t}=(P_{t}^{\mathrm{L}},P_{t}^{\mathrm{PV}},E_{t})$. Let $s_{\mathrm{lp},t}=(P_{t}^{\mathrm{L}},P_{t}^{\mathrm{PV}})$ and $s_{\mathrm{e},t}=E_{t}$, we have $s_{t}=(s_{\mathrm{lp},t},s_{\mathrm{e},t})$.\par
    
    Due to the uncertainties of future load and renewable power generation, the agent is unable to observe $P_{t}^{\mathrm{L}}$ and $P_{t}^{\mathrm{PV}}$ at the beginning of time step $t$. Instead, it receives observation $o_{t}$ of the system at the beginning of time step $t$, where we define the observation as $o_{t}=(P_{t-1}^{\mathrm{L}},P_{t-1}^{\mathrm{PV}},E_{t})$. Note that $P_{t-1}^{\mathrm{L}}$ and $P_{t-1}^{\mathrm{PV}}$ are the aggregated power demand of loads and power output of PV at time step $t-1$, respectively, which are readily available to the agent at the beginning of time step $t$. Let $o_{\mathrm{lp},t}=(P_{t-1}^{\mathrm{L}},P_{t-1}^{\mathrm{PV}})$ and we have $o_{t}=(o_{\mathrm{lp},t},s_{\mathrm{e},t})$. Note that $o_{\mathrm{lp},t}=s_{\mathrm{lp},t-1}$.  \par
    
    The state space and observation space $\mathcal{S}=\mathcal{O}=[P_{\mathrm{min}}^{\mathrm{L}},P_{\mathrm{max}}^{\mathrm{L}}]\times[P_{\mathrm{min}}^{\mathrm{PV}},P_{\mathrm{max}}^{\mathrm{PV}}]\times[E_{\mathrm{min}},E_{\mathrm{max}}]$, where $P_{\mathrm{min}}^{\mathrm{L}}$ and $P_{\mathrm{max}}^{\mathrm{L}}$ are the minimum and maximum aggregated loads, while $P_{\mathrm{min}}^{\mathrm{PV}}$ and $P_{\mathrm{max}}^{\mathrm{PV}}$ are the minimum and maximum aggregated power outputs of PV.   \par
    
    \subsection{Action}
    We define the action at time step $t$, $\forall t\in\{1,2,\cdots,T\}$, as $a_{t}=\{P_{t}^{\mathrm{DG}_{d}}\}_{d=1}^{D}$, which are the output power of DGs. The action space $\mathcal{A}\in\bigcup_{d=1}^{D}[P_{\mathrm{min}}^{DG_{d}},P_{\mathrm{max}}^{DG_{d}}]$. \par
    
    \subsection{Policy}
    At time step $t$, the agent requires access to state $s_{t}$ to determine the optimal action $a_{t}$, i.e., $a_{t}=\mu_{t}(s_{t})$, where $\mu_{t}$ is a decision rule that prescribes a procedure for action selection at time step $t\in\{1,\cdots,T\}$. The sequence of decision rules for all the time steps forms a policy $\pi$, i.e., $\pi=(\mu_{1},\cdots,\mu_{T})$. In the rest of this paper, we will refer to $\mu_{t}, \forall t\in{1,\cdots,T}$ as a policy or policy function instead of a decision rule with a slight abuse of terminology. \par 
    
    \newtheorem{remark}{Remark}
	\begin{remark}[\textbf{Stationary and Non-Stationary Policy}]	
    A policy is stationary if $\mu_{t}=\mu$ for all $t\in\{1,\cdots,T\}$, which are fundamental to the theory of infinite-horizon MDP \cite{Puterman2014}. On the other hand, the optimal policy for a finite-horizon MDP as studied in this paper is normally non-stationary \cite{Grondman2013}. This is because the value functions for the same system state are different at different time steps. 
	\end{remark}    
    
    \subsection{History}
    As only the observation $o_{t}$ is available instead of $s_{t}$, the agent is provided with history $h_{t}$ to derive action $a_{t}=\mu_{t}(h_{t})$. \par 
    
    In a POMDP problem, the history is normally defined as $h_{t}=(o_{1},a_{1},o_{2},a_{2},\cdots,o_{t-1},a_{t-1},o_{t})$. In this paper, the history provided to the agent is tailored for the MG energy management problem as below.

    \newtheorem{definition}{Definition}
    \begin{definition}[\textbf{Definition of history}]	
     We define history $h_{t}=(o_{\mathrm{lp},t-\tau+1},\cdots,o_{\mathrm{lp},t-1},o_{t})$, $\forall t\in\{1,\cdots,T\}$, where $\tau$ is the size of the past observation window.
    \end{definition}

    Note that compared with the standard history definition, our definition does not include action history $\{a_{t'}\}_{t'=t-\tau}^{t-1}$ nor battery SoC history $\{s_{\mathrm{e},t'}\}_{t'=t-\tau}^{t-1}$ up to time step $t-1$. This is because we only need the historical load and PV generation data $\{o_{\mathrm{lp},t'}\}_{t'=t-\tau+1}^{t}$ up to time step $t-1$ to predict the load and PV state $s_{\mathrm{lp},t}$ at time step $t$. \par
    
    The evolution of history can be derived as
    \begin{equation}  
    \label{eq9} 
    h_{t+1}=h_{t}\setminus \{s_{\mathrm{e},t}\}\cup \{o_{t+1}\}
    \end{equation}
    \subsection{Reward Function}
    The optimization objective is to minimize the total power generation cost and power unbalance in the MG within a $24$-hour time horizon. Therefore, we define the reward function as 
    \begin{equation}
    \label{eq10}	
    r_{t}(s_{t},a_{t})=-(k_{1}\sum_{d=1}^{D}c_{t}^{\mathrm{DG}_{d}}+k_{2}c_{t}^{\mathrm{US}}),
    \end{equation}
    \noindent where $k_{1}$ and $k_{2}$ are weights indicating the relative importance of minimizing the power generation cost versus power unbalance. $c_{t}^{\mathrm{DG}_{d}}$ is given in \eqref{eq2}, while $c_{t}^{\mathrm{US}}$ is the cost of the aggregated unserved or wasted active power, i.e.,
    
    \begin{equation}
    \label{eq11}	
    c_{t}^{\mathrm{US}}=\left\{
    \begin{array}{ll}
    k_{21}(\delta_{t}-P_{\mathrm{ch\_lim}}^{\mathrm{E}})\Delta t, & \mathrm{if} \ \delta_{t}>P_{\mathrm{ch\_lim}}^{\mathrm{E}}\\
    -k_{22}(\delta_{t}+P_{\mathrm{dis\_lim}}^{\mathrm{E}})\Delta t, & \mathrm{if} \ \delta_{t}<-P_{\mathrm{dis\_lim}}^{\mathrm{E}} \\
    0, & \mathrm{otherwise} \\
    \end{array}\right. ,
    \end{equation}
    \noindent where $\delta_{t}$, $P_{\mathrm{ch\_lim}}^{\mathrm{E}}$, and $P_{\mathrm{dis\_lim}}^{\mathrm{E}}$ are given in \eqref{eq4}, \eqref{eq7}, and \eqref{eq8}, respectively. $k_{21}$ and $k_{22}$ are weights that indicate the relative importance of avoiding unserved power situation and wasted power situation.

    \subsection{Transition Probability}
    The state transition probability $\mathrm{Pr.}(s_{t+1}|s_{t},a_{t})$ can be derived by
    \begin{align}
	\label{eq12}
	&\mathrm{Pr}(s_{t+1}|s_{t},a_{t})= \IEEEnonumber \\ 
	&\mathrm{Pr}(P_{t+1}^{\mathrm{L}}|P_{t}^{\mathrm{L}})\mathrm{Pr}(P_{t+1}^{\mathrm{PV}}|P_{t}^{\mathrm{PV}})\mathrm{Pr}(E_{t+1}|E_{t},a_{t}),
	\end{align}	    
    
    \noindent where $\mathrm{Pr}(E_{t+1}|E_{t},a_{t})$ is given in \eqref{eq3}. However, the transition probabilities of the load power demands and PV power outputs, i.e.,  $\mathrm{Pr}(P_{t+1}^{\mathrm{L}}|P_{t}^{\mathrm{L}})$ and $\mathrm{Pr}(P_{t+1}^{\mathrm{PV}}|P_{t}^{\mathrm{PV}})$ are not available. We will use model-free RL algorithms to learn the solution to the above POMDP problem based on real-world data.\par

    \subsection{Model Formulation}
  
   Define the \emph{return} from a state as the sum of rewards over the finite horizon starting from that state. Given the above elements, if state $s_{t}$ is considered to be known at each time step $t$, the return is given as
   \begin{equation}
   	\label{eq13}
   R^{\mathrm{MDP},\pi}=\sum_{t=1}^{T}r_{t}(s_{t},\mu_{t}(s_{t})).
   \end{equation}
   
   	On the other hand, if only observation $o_{t}$ instead of state $s_{t}$ is considered to be known at each time step $t$, which is the practical case, the return is given as
   \begin{equation}
   	\label{eq14}
    R^{\mathrm{POMDP,\pi}}=\sum_{t=1}^{T}r_{t}(s_{t},\mu_{t}(h_{t})).
   \end{equation}   	
   
    Therefore, a finite-horizon MDP model or POMDP model can be formulated as
    \begin{equation}
	\label{eq15}
	\max_{\pi} \mathrm{E}[R^{\pi}],
	\end{equation}    
	\noindent where $R^{\pi}$ represents $R^{\mathrm{MDP},\pi}$ or $R^{\mathrm{POMDP},\pi}$, respectively. Note that the expectations of the returns in \eqref{eq13} and \eqref{eq14} are taken w.r.t. the
	unique steady-state distribution induced by the given policy $\pi=(\mu_{1},\cdots,\mu_{T})$.\par
	
    
    \newtheorem{remark2}[remark]{Remark}
    \begin{remark2}[\textbf{Terminal Reward}]	
    In a finite-horizon MDP, the reward $r_{T}$ for the last time step is referred to as the terminal reward, which is normally considered to be independent of the action $a_{T}$. In the MDP and POMDP models formulated in this paper, the terminal reward is still dependent on action $a_{T}$, where the optimal action can be derived by a \emph{myopic policy}, i.e., $\mu_{T}=\mu^{\mathrm{mo}}$ that optimizes the terminal reward $r_{T}$ without having to consider any future reward. \par 
    
 A \emph{myopic policy} optimizes the reward function $r_{t}(s_{t},a_{t})$ of a MDP at each time step without considering the impact of action $a_{t}$ on the future rewards. Note that for the MDP model, the myopic policy $\mu^{\mathrm{mo}}$ can be derived directly by minimizing \eqref{eq10} without learning. However, this is not the case for the POMDP model as the state information is not available.\par
    \end{remark2}    
	
	\section{DRL Algorithms}

	\subsection{Algorithm for Finite-Horizon MDP: FH-DDPG}
	DRL algorithms are known to be unstable, and there are many challenges when applied to solve real-world problems \cite{rlblogpost}. Moreover, it is usually applied to solve infinite-horizon MDP problems. In this paper, we design a DRL algorithm for the formulated finite-horizon MDP model based on a baseline algorithm, i.e., DDPG. The instability problem and the unique characteristics of the finite-horizon setting are addressed using two key ideas as discussed below.\par
	
    Firstly, in order to capture the time-varying policy $\pi=(\mu_{1},\cdots,\mu_{T})$ under a finite-horizon setting, we train $T-1$ actor networks $\{\mu_{t}(s|\theta^{\mu_{t}})\}_{t=1}^{T-1}$ to approximate policy functions $\{\mu_{t}(s)\}_{t=1}^{T-1}$, instead of using only one actor network as in most up-to-date actor-critic algorithms \cite{lillicrap2015,mnih2016,schulman2015,heess2015}. The policy $\mu_{T}$ at the last time step $T$ can be directly set to be the same as the myopic policy $\mu^{\mathrm{mo}}$. \par 
    
    Secondly, in order to address the instability problem and achieve faster convergence, we use backward induction and divide the whole training process into training $T-1$ one-period MDPs in backward sequence, starting from time step $T-1$ and going backward to time step $1$. In training the one-period MDP for each time step $t\in\{T-1,\cdots,1\}$, DDPG is applied to derive the actor $\mu_{t}(s|\theta^{\mu_{t}})$ along with a critic $Q(s,a|\theta^{Q})$. Then, the actor weights are stored, and both the actor and critic trained for time step $t$ are used as the target actor and critic for the previous time step $t-1$, which is trained next in the proposed algorithm. In this way, we start by having the optimal policy for a single time step $T$, i.e., $\pi=(\mu_{T}(s)=\mu^{\mathrm{mo}}(s))$ before training. And then, we derive the optimal policy starting from time step $T-1$ until the end of the time horizon after we train the one-period MDP for time step $T-1$, i.e., $\pi=(\mu_{T-1}(s|\theta^{\mu_{T-1}}),\mu_{T}(s))$. This process keeps going on until the training of the one-period MDP for time step $1$ is finished, which gives us the complete optimal policy starting from time step $1$ until the end of the time horizon, i.e., $\pi=(\mu_{1}(s|\theta^{\mu_{1}}),\cdots\mu_{T-1}(s|\theta^{\mu_{T-1}}),\mu_{T}(s))$. Note that DDPG is always used to solve the one-period MDPs in which an episode only consists of two time steps, and only the actor and critic of the first time step need to be trained. By greatly reducing the number of time steps within an episode, the performance of DDPG proves to be much more stable. \par
    
    The proposed DRL algorithm for the finite-horizon MDP, namely FH-DDPG, is given in Algorithm 1, where $\alpha$ and $\beta$ denote the learning rates for the actor and critic networks, respectively. Note that the neural network architectures of the actor and critic in the FH-DDPG algorithm are the same with those in the DDPG algorithm \cite{lillicrap2015}.\par
	
	\begin{algorithm}
		\caption{FH-DDPG Algorithm}
		\label{alg1}
		\begin{algorithmic}
			\STATE Randomly initialize actor network $\mu(s|\theta^{\mu})$ and critic network $Q(s,a|\theta^{Q})$ with weights $\theta^{\mu}=\theta^{\mu0}$ and $\theta^{Q}=\theta^{Q0}$ 
		    Initialize target networks $Q'$ and $\mu'$ with $\theta^{Q'}\leftarrow\theta^{Q}$ and $\theta^{\mu'}\leftarrow\theta^{\mu}$
			\FOR{$t= T - 1, \cdots, 1$}
			\STATE{Initialize replay buffer $R$}
			\STATE{Initialize a random process $\mathcal{N}$ for action exploration}
			\FOR{episode $e = 1,\dots, M$ }
			\STATE{Receive state $s_{t}^{(e)}$}
			\STATE{Select action $a_{t}^{(e)}$ according to the current policy and exploration noise}
			\STATE{Execute action $a_{t}^{(e)}$ and observe reward $r_{t}^{(e)}$ and observe new state $s_{t+1}^{(e)}$}
			\STATE{Store transition $(s_{t}^{(e)},a_{t}^{(e)},r_{t}^{(e)},s_{t+1}^{(e)})$ in $R$}
			\STATE{Sample a random minibatch of $N$ transitions $(s_{t}^{(i)},a_{t}^{(i)},r_{t}^{(i)},s_{t+1}^{(i)})$ from $R$}
			\IF{$t = T-1$}
			\STATE{Set $y_{t}^{(i)}=r_{t}^{(i)}+\gamma r_{T} (s_{t + 1}^{(i)},\mu^{\mathrm{mo}} (s_{t + 1}^{(i)}))$}
			\ELSE
			\STATE{Set $y_{t}^{(i)}=r_{t}^{(i)}+\gamma Q'(s_{t + 1}^{(i)},\mu'(s_{t + 1}^{(i)}|\theta^{\mu'})|\theta^{Q'})$}
			\ENDIF
			\STATE Update critic by minimizing the loss: 
			\begin{displaymath}
				L=\frac{1}{N}\sum_{i} (y_{t}^{(i)}-Q(s_{t}^{(i)},a_{t}^{(i)} |\theta^{Q}))
			\end{displaymath}
		
			\begin{displaymath}
			\theta^{Q}\leftarrow\theta^{Q}+\beta \bigtriangledown_{\theta^{Q}}L
			\end{displaymath}
		
			\STATE Update the actor using the sampled policy gradient:
			\begin{align}
			 & \bigtriangledown_{\theta^{\mu}}J\approx \IEEEnonumber \\ 
			 &\frac{1}{N}(\sum_{i}\bigtriangledown_{a}Q(s,a|\theta^{Q})|_{s=s_{t}^{(i)},a=\mu(s_{t}^{(i)})}\bigtriangledown_{\theta^{\mu}}\mu(s|\theta^{\mu})|_{s_{t}^{(i)}}) \IEEEnonumber
			\end{align}
		
				\begin{displaymath}
				\theta^{\mu}\leftarrow\theta^{\mu}+\alpha \bigtriangledown_{\theta^{\mu}}J
				\end{displaymath}
		
			\ENDFOR
			\STATE Update the target network:
			\begin{displaymath}
			\theta^{Q'}\leftarrow\theta^{Q}, \ \theta^{\mu'}\leftarrow\theta^{\mu}
			\end{displaymath}

			\STATE Save weight of actor network:
			\begin{displaymath}
			\theta^{\mu_{t}}\leftarrow\theta^{\mu}
			\end{displaymath}
			\STATE Reset weight of actor and critic networks to initial value:
			\begin{displaymath}
			\theta^{Q}\leftarrow\theta^{Q0}, \  \theta^{\mu}\leftarrow\theta^{\mu0}
			\end{displaymath}
			\ENDFOR
			\end{algorithmic}
	\end{algorithm}
	
    \subsection{Algorithm for Finite-Horizon POMDP: FH-RDPG}
    The above FH-DDPG algorithm is designed to solve finite-horizon MDP problems. However in practice, the state information cannot be obtained at each time step. Therefore, in order to address the partial observable problem for finite-horizon POMDP, we propose another algorithm, namely FH-RDPG. In general, it is harder to use backward induction for solving POMDP as the history needs to be obtained first from Monte-Carlo simulation. However, in the MG environment under consideration in this paper, we can tackle this challenge by defining the history as given in Definition 1. Specifically, we only need the history data of the load and PV, i.e., $\{o_{\mathrm{lp},t'}\}_{t'=t-\tau}^{t}$ up to time step $t-1$,  which can be obtained from real-world data without any need for simulation, to predict the load and PV state $s_{\mathrm{lp},t}$ at time step $t$. On the other hand, the history action and battery SoC statistics that require simulation to obtain samples of trajectories are not needed to derive the optimal action for time step $t$. \par  
    
    The FH-RDPG Algorithm is given in Algorithm 2. Note that it is similar to the FH-DDPG algorithm except that state $s_{t}$ is replaced with history $h_{t}$. Moreover, as the optimal policy for the last time step $T$ cannot be obtained directly as in the case of FH-DDPG due to the lack of state information, we need to learn the optimal policy $\mu_{T}(h_{t})$ by training the actor and critic as well. As a result, a total of $T$ actors are obtained. \par
    
\begin{figure}[!htb]
    	\centering
    	\includegraphics[width=0.48\textwidth]{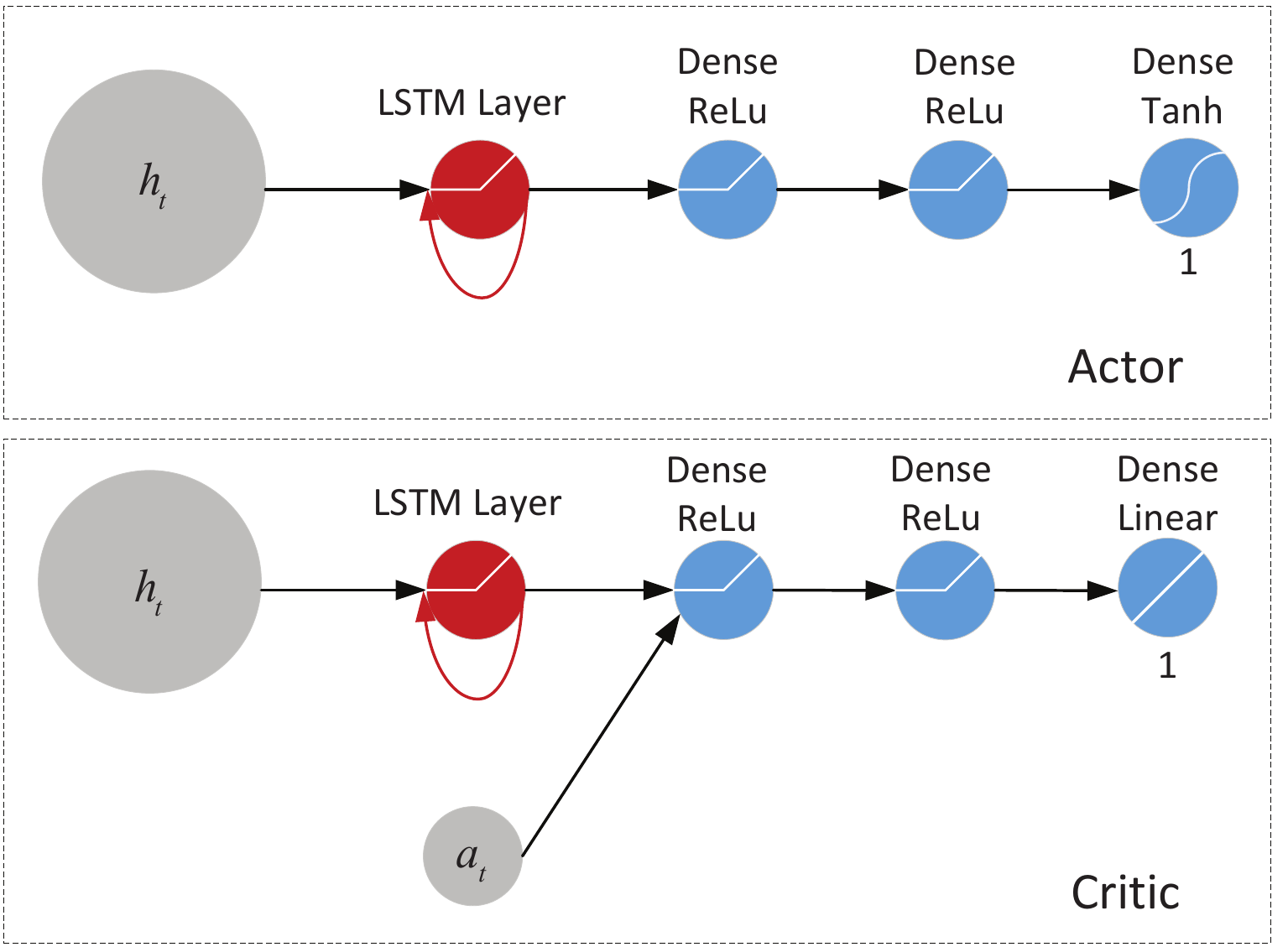}
    	\caption{Actor and Critic Network Architecture of FH-RDPG Algorithm.}
    	\label{FHRDPGArch}
    \end{figure}

    Theoretically, we can use the feed-forward neural networks in DDPG or FH-DDPG for the FH-RDPG algorithm as well. However, it is well-known that RNNs can learn to preserve information about the past using backpropagation through time (BPTT). Moreover, Long Short-Term Memory Networks (LSTMs) are a type of widely used RNNs that are able to solve shortcomings in the RNN, e,g., vanishing gradient, exploding gradient and long term dependencies, etc.. Therefore, we replace the first feed-forward layers in both the actor and critic networks in FH-DDPG with LSTM layers in the FH-RDPG algorithm. The neural network architectures of the actor and critic in FH-RDPG algorithm are given in Fig. \ref{FHRDPGArch}.

    	\begin{algorithm}
    	\caption{FH-RDPG Algorithm}
    	\label{alg2}
    	\begin{algorithmic}
    		\STATE Randomly initialize actor network $\mu(s|\theta^{\mu})$ and critic network $Q(s,a|\theta^{Q})$ with weights $\theta^{\mu}=\theta^{\mu0}$ and $\theta^{Q}=\theta^{Q0}$ 
    		Initialize target networks $Q'$ and $\mu'$ with $\theta^{Q'}\leftarrow\theta^{Q}$ and $\theta^{\mu'}\leftarrow\theta^{\mu}$
      		\FOR{$t=T,\cdots,1$}
    		\STATE{Initialize replay buffer $R$}
    		\STATE{Initialize a random process $\mathcal{N}$ for action exploration}
    		\FOR{episode $e = 1,\dots, M$ }
    		\STATE{Receive history $h_{t}^{(e)}$}
    		\STATE{Select action $a_{t}^{(e)}$ according to the current policy and exploration noise}
    		\STATE{Execute action $a_{t}^{(e)}$ and observe reward $r_{t}^{(e)}$ and observe new observation $o_{t+1}^{(e)}$}
    		\STATE{Store transition $(h_{t}^{(e)},a_{t}^{(e)},r_{t}^{(e)},o_{t+1}^{(e)})$ in $R$}
    		\STATE{Sample a random minibatch of $N$ transitions $(h_{t}^{(i)},a_{t}^{(i)},r_{t}^{(i)},o_{t+1}^{(i)})$ from $R$}
    		\IF{$t = T$}
    		\STATE{Set $y_{T}^{(i)}=r_{T}^{(i)}$}
    		\ELSE
    		\STATE{Construct $h_{t+1}$ by \eqref{eq9}}
    		\STATE{Set $y_{t}^{(i)}=r_{t}^{(i)}+\gamma Q'(h_{t + 1}^{(i)},\mu'(h_{t + 1}^{(i)}|\theta^{\mu'})|\theta^{Q'})$}
    		\ENDIF
    		\STATE Update critic by minimizing the loss: 
    		\begin{displaymath}
    		L=\frac{1}{N}\sum_{i} (y_{t}^{(i)}-Q(h_{t}^{(i)},a_{t}^{(i)} |\theta^{Q}))
    		\end{displaymath}
    	
    			\begin{displaymath}
    			\theta^{Q}\leftarrow\theta^{Q}+\beta \bigtriangledown_{\theta^{Q}}L
    			\end{displaymath}
    	
    		\STATE Update the actor using the sampled policy gradient:
    		\begin{align}
    		& \bigtriangledown_{\theta^{\mu}}J\approx \IEEEnonumber \\ &\frac{1}{N}(\sum_{i}\bigtriangledown_{a}Q(h,a|\theta^{Q})|_{h=h_{t}^{(i)},a=\mu(h_{t}^{(i)})}\bigtriangledown_{\theta^{\mu}}\mu(h|\theta^{\mu})|_{h_{t}^{(i)}}) \IEEEnonumber
    		\end{align}
    	
    			\begin{displaymath}
    			\theta^{\mu}\leftarrow\theta^{\mu}+\alpha \bigtriangledown_{\theta^{\mu}}J
    			\end{displaymath}
    	
    		\ENDFOR
    		\STATE Update the target network:
    		\begin{displaymath}
    		\theta^{Q'}\leftarrow\theta^{Q}, \ \theta^{\mu'}\leftarrow\theta^{\mu}
    		\end{displaymath}
  		
    		\STATE Save weight of actor network:
    		\begin{displaymath}
    		\theta^{\mu_{t}}\leftarrow\theta^{\mu}
    		\end{displaymath}
    		\STATE Reset weight of actor and critic networks to initial value:
    		\begin{displaymath}
    		\theta^{Q}\leftarrow\theta^{Q0}, \ \theta^{\mu}\leftarrow\theta^{\mu0}
    		\end{displaymath}
    		\ENDFOR
    	\end{algorithmic}
    	\end{algorithm}
    
    \subsection{Training of the DRL Algorithms}
    In order to obtain the best performance for a specific day, the DRL algorithms should be trained based on the load and PV data on that day. The obtained policy will be trained to best suit the statistics of that particular day.\par
    
    However in reality, the load and PV data for the future is not available to train the DRL algorithms. There are generally two methods to address this problem. The first method involves predicting the load and PV data for the future based on the past data, and then train the DRL algorithms using the predicted data. However, despite the vast amount of existing research for PV and load prediction, the predicted data is still subject to inaccuracies and errors. Therefore, the performance of the first method depends largely on the accuracy of the prediction algorithms. \par
    
    In the second method, the policy is obtained by training DRL algorithms based on data from the past, and it will be applied for making energy management decisions in the future, without knowing the future PV and load data. The advantage of this method is that it learns the policy in one go without depending on the prediction algorithms.   \par 
    
    \newtheorem{remark3}[remark]{Remark}
\begin{remark3}[\textbf{Two levels of uncertainties}]	
In our proposed DRL model, the impact of uncertainty in PV generation and load demand on MG performance is decoupled into two levels. The first level is the ``uncertainty for the next time step", which is captured by the POMDP model. Specifically, the agent can only obtain observation $o_{t}=(P_{t-1}^{\mathrm{L}},P_{t-1}^{\mathrm{PV}},E_{t})$ instead of state $s_{t}=(P_{t}^{\mathrm{L}},P_{t}^{\mathrm{PV}},E_{t})$ at the beginning of time step $t$. The second level is the ``uncertainty for the next day", where we train the DRL algorithms using history data to make decisions in the future, where the PV and load statistics are unknown. With the proposed DRL model, we are able to evaluate the impact of both levels of uncertainty on the MG performance in Section V: (1) compare the performance between the MDP and POMDP models to assess the impact of ``uncertainty for the next time step"; (2) compare the performance of training and evaluation based on the same day data versus different days data to assess the impact of ``uncertainty for the next day".
\end{remark3}

    \section{Experimental Results}
In this section, we present the simulation results of the proposed FH-DDPG and FH-RDPG algorithms as well as several benchmark DRL and non-DRL algorithms. All the algorithms are trained/tested using the real isolated MGs data from Ergon Energy's $264$ kilowatt solar farm at remote Doomadgee in north-west Queensland. The MG environment and the algorithms are implemented in Tensorflow 1.13 using Python, which is an open source DL platform \cite{abadi2016tensorflow}.\par

 \subsection{Benchmark Algorithms}
As discussed in Remark 3, the impact of uncertainty on MG performance can be decoupled into two levels. In order to evaluate the performance of the proposed algorithms under different levels of uncertainties, we designed four cases with benchmark algorithms implemented in each case as follows.

\subsubsection{Case I: MDP, train and test with same-day data} 
In MDP environment, the load and PV data for time step $t$ are considered to be known at the beginning of the time step, which eliminates the ``uncertainty for the next time step". When the DRL algorithms are trained and tested based on data for the same day, the PV and load for the finite horizon is assumed to be known according to the given data, and the ``uncertainty for the next day" is eliminated. Therefore, there is no uncertainty in Case I, where the energy dispatch problem becomes a \emph{fully-observable deterministic sequential decision problem}. \par 

The FH-DDPG algorithm is implemented in Case I along with the following benchmark algorithms:
\begin{itemize}
	\item DDPG algorithm, which is a baseline DRL algorithm;
	\item myopic algorithm, where the action $a_{t}$ of each time step $t\in\{1,\cdots,T\}$ is directly derived by minimizing the reward function $r_{t}(s_{t},a_{t})$ in (10) without considering the impact of action $a_{t}$ on the future rewards;
    \item Iterative Linear Quadratic Gaussian (iLQG) algorithm \cite{Todorov2005}, which is a well-known model-based planner used for optimal control in nonlinear system with continuous states and actions. 
\end{itemize}   

\subsubsection{Case II: POMDP, train and test with same-day data} 
In POMDP environment, the history of load and PV data up to time step $t-1$ instead of the data for time step $t$ are considered to be known at the beginning of time step $t$. Therefore, Case II considers the ``uncertainty for the next time step", but not ``uncertainty for the next day". The energy dispatch problem becomes a \emph{partially-observable deterministic sequential decision problem}. \par

The FH-RDPG algorithm is implemented in Case II along with the following benchmark algorithms:
\begin{itemize}
	\item RDPG algorithm, which is a baseline DRL algorithm;
	\item myopic-POMDP algorithm, which uses observation $o_{t}$ in place of $s_{t}$ in (10) to directly derive the action as $\arg\min_{a_{t}}r_{t}(o_{t},a_{t})$;
	\item iLQG-POMDP algorithm, which is similar to iLQG except that state $s_{t}$ is replaced by observation $o_{t}$. 
\end{itemize}   

\subsubsection{Case III: MDP, train with history data} 
The DRL algorithms are trained with history data instead of the same-day data used for testing. Therefore, Case III considers the ``uncertainty for the next day", but not ``uncertainty for the next time step". The energy dispatch problem becomes a \emph{fully-observable stochastic sequential decision problem}. \par 

The FH-DDPG algorithm is implemented in Case III along with the following benchmark algorithms:
\begin{itemize}
	\item DDPG algorithm, which is trained using history data;
	\item MPC-iLQG algorithm \cite{lillicrap2015}, which iteratively applies iLQG at each time step with predicted PV and load for the future horizon using RNN. 
\end{itemize}   

\subsubsection{Case IV: POMDP, train with history data} 
Case IV considers both the ``uncertainty for the next day" and ``uncertainty for the next time step". The energy dispatch problem becomes a \emph{partially-observable stochastic sequential decision problem}. \par 

The FH-RDPG algorithm is implemented in Case IV along with the following benchmark algorithms:
\begin{itemize}
	\item RDPG algorithm, which is trained using history data;
	\item MPC-iLQG-POMDP algorithm, which iteratively applies iLQG-POMDP at each time step with predicted PV and load for the future horizon using RNN. 
\end{itemize}

 \subsection{Experiment Setup}
The technical constraints and operational parameters of the MG are given in Table \ref{env_para}. The parameters involved in the battery model include the maximum charging/discharging power of the battery, the maximum and minimum SoC and charging/discharging efficiency \cite{Zeng2018}. The operational parameters of the DGs are also summarized in Table \ref{env_para} \cite{djurovic2012simplified}.  The number of DGs $D$ is set to be $1$ in accordance with the configuration at Doomadgee. It is straightforward to apply our proposed algorithms to scenarios with multiple DGs. When there are $N$ DGs, the sizes of the state space for FH-DDPG and FH-RDPG are $3^{N}$ and $9^{N}$, respectively, while the size of the action space is $N$. As the number of DGs in an isolated MG is usually small, the sizes of the state and action spaces are acceptable. Due to the space limitation of the paper, we do not include results for $D>1$.\par

The interval for each time step is set to $\Delta t =1$ hour. As there are $T=24$ time steps in one day, each episode is comprised of 24 time steps. In each time step $t$, the agents in FH-DDPG and DDPG can receive the PV and load data $s_{\mathrm{lp},t}$ of the current time step. On the other hand, only the history of PV and load data $\{o_{\mathrm{lp},t'}\}_{t'=t-\tau}^{t}$ up to time step $t-1$ is available to the agents in FH-RDPG and RDPG. We set the history window $\tau=4$ in our simulation, as this value achieves the best performance. \par

Without explicit specification, the coefficients in the reward function as given in \eqref{eq10} are set to $k_{1}=0.001$ and $k_{2}=1$, respectively. Moreover, we set $k_{21}=k_{22}=1$ in \eqref{eq11}. The impact of the coefficients on performance are evaluated in Section V.C.5).\par

\begin{table}[!htb]
	\renewcommand{\arraystretch}{1.3}
	\caption{Technical constraints and operational parameters of the MG} \label{env_para} \centering
	\begin{tabular}{cccccc}
		\hline
		\multirow{3}*{\textbf{Battery}} & \multirow{2}*{$P_{\mathrm{max}}^{\mathrm{E}}$} & \multirow{2}*{$E_{\mathrm{max}}$} & \multirow{2}*{$E_{\mathrm{min}}$} & \multirow{2}*{$\eta_{\mathrm{ch}}$} & \multirow{2}*{$\eta_{\mathrm{dis}}$} \\
		\\
		\cline{2-6} 
		& 120kW &  2000kWh & 24kWh  & 0.98  &0.98 \\
		\hline
		 &﻿\multirow{2}*{$P_{\mathrm{max}}^{DG_{d}}$}  &﻿\multirow{2}*{$P_{\mathrm{min}}^{DG_{d}}$}
& \multicolumn{3}{c} {coefficients of cost curve} \\
		\cline{4-6}
		\textbf{DG}&&& $a_{d}$ &﻿$b_{d}$ &﻿$c_{d}$ \\
		\cline{2-6} 
	 &600kW& 100kW& 0.005& 6 &100 \\
		\hline
	\end{tabular}
\end{table}  

The hyper-parameters for training are summarized in Table \ref{alg_para}. The values of all the hyper-parameters were selected by performing a grid search as in \cite{Mnih2015}, using the hyper-parameters reported in \cite{lillicrap2015} as a reference. The network architecture and activation functions of the FH-RDPG and RDPG algorithms are given in Fig. \ref{FHRDPGArch}, while those of the FH-DDPG and DDPG algorithms are similar except that the LSTM layers are replaced by the Dense layers with ReLu activation function. The sizes of the neural networks in the simulation are given in Table \ref{alg_para}. For example, there are three hidden layers in the actor and critic networks of FH-DDPG, where the number of neurons in each layer is $400$, $300$, and $100$, respectively. Note that the size of input layer for the actor of FH-DDPG is decided by the dimension of state $s_t$, which is $3$. Meanwhile, the size of input layer for the actor of FH-RDPG is decided by the dimension of history $h_t$ given in Definition 1, which is $2\times\tau+1$ with $\tau=4$ being the length of history window as given in Table \ref{alg_para}. For the critics of both FH-DDPG and FH-RDPG, an additional $1$-dimensional action input is fed to the second hidden layer. The size of replay buffer is set to be $20000$ in all the experiments. When the replay buffer is full, the oldest sample will be discarded before a new sample is stored into the buffer. \par


\begin{table*}[!htb]
	\renewcommand{\arraystretch}{1.3}
	\caption{Hyper-Parameters of the DRL algorithms for training} \label{alg_para} \centering
	\begin{tabular}{ccccc}
		\hline
		{\textbf{Parameter}} & \multicolumn{4}{c} {\textbf{Value}} \\
		\hline
		& FH-DDPG &FH-RDPG&DDPG&RDPG\\
		\hline
		Actor network size &$400,300,100$& $128,128,64$ & $256,128$& $128,128$\\
		\hline
		Critic network size & $400,300,100$ & $128,128,64$ & $256,128$& $128,128,256$\\
		\hline
		Actor learning rate $\alpha$ & $5$e$-6$ &$5$e$-6$ & $1$e$-6$ & $1$e$-6$\\
		\hline
		Critic learning rate $\beta$ & $5$e$-5$ &$5$e$-5$ & $1$e$-5$&$1$e$-5$\\
		\hline
		History window $\tau$ & / &4 & / & $4$ \\
		\hline
		Replay buffer size &  \multicolumn{4}{c} {$20000$} \\
		\hline
		Batch size & \multicolumn{4}{c} {$128$} \\
		\hline
		Reward scale & \multicolumn{4}{c} {$2$e$-3$}  \\
		\hline
	    Soft target update & /  &  /  &  $0.001$& $0.001$  \\
		\hline
		Noise type & \multicolumn{4}{c}{Ornstein-Uhlenbeck Process with $\theta=0.15$ and $\sigma=0.5$} \\
		\hline
		Final layer weights/biases initialization & \multicolumn{4}{c}{Random uniform distribution $[-3\times10^{-3},3\times10^{-3}]$} \\
		\hline
	\end{tabular}
\end{table*}

     \subsection{Train and test based on same-day data (Case I and Case II)}
    We first train and test the DRL algorithms based on the load and PV data for the same day, i.e., Case I and Case II. The PV and load data from 23:59 PM on July 07, 2017 to 22:59 PM on July 08, 2017 are used to both train and test the algorithms. The SoC at the beginning of the day is initialized to be a random value. The purpose of the DRL algorithms is to learn the best energy management policy for one specific day from the data. In this case, we can focus on the comparison between the learning capabilities of different DRL algorithms without being affected by the noises due to the statistics discrepancies in different days.  \par  
    
    Table \ref{perform} summarizes the performance of two baseline DRL algorithms - DDPG and RDPG, and the performance of our proposed two DRL algorithms - FH-DDPG and FH-RDPG. In addition to the four DRL algorithms, we also report the performance of the myopic algorithm and iLQG algorithm in both the MDP and POMDP environments.  \par

    
    \begin{table*}[!htb]
	\renewcommand{\arraystretch}{1.3}
	\caption{Performance after training across 5 different runs. Each run has 24 steps in total. We report both the average and best observed performance (across 5 runs). We present the individual performance for each run for DDPG, FH-DDPG, RDPG, and FH-RDPG algorithms. For comparision we also include results from myopic, myopic-POMDP, iLQG, and iLQG-POMDP algorithms. } \label{perform} \centering
	\begin{tabular}{|c|c|c|c|c|c|c|c|c|c|}
		\hline
		\textbf{Environment} & \textbf{Algorithm} & \multicolumn{8}{|c|}{\textbf{Performance}}  \\
		\cline{3-10}
		& & \textbf{Run 1} & \textbf{Run 2}& \textbf{Run 3} & \textbf{Run 4} & \textbf{Run 5} &\textbf{Max} & \textbf{Average} & \textbf{Std Error} \\
		\hline
		\textbf{MDP}  	& \textbf{myopic} & \multicolumn{7}{|c|}{$-1.1488$}   & $0$\\
		\cline{2-10} 
		& \textbf{iLQG} & \multicolumn{7}{|c|}{$-0.3003$} & $0$ \\
		\cline{2-10}
		& \textbf{DDPG} & $-0.2817$ & $-0.8986$ & $-8.5000$ & $-1.4037$ & $-0.7835$ &  $-0.2817$ & $-2.3734$  & $3.4477$\\
		\cline{2-10}
		& \textbf{FH-DDPG} & $-0.2312$ & $-0.2308$ & $-0.2310$ & $-0.2315$ & $-0.2438$  & $-0.2308$ & $-0.2336$ & $0.0057$ \\		
		\hline
		\textbf{POMDP}  & \textbf{myopic-POMDP} &\multicolumn{7}{|c|}{$-1.5082$}  & $0$\\
		\cline{2-10}
		& \textbf{iLQG-POMDP} & \multicolumn{7}{|c|}{$-0.3541$} & $0$ \\
		\cline{2-10}
		& \textbf{RDPG} & $-0.3887$  & $-0.4344$ & $-0.4474$ & $-0.4179$ & $-0.3299$ & $-0.3299$ & $-0.4036$ & $0.0467$ \\ 
		\cline{2-10}
		& \textbf{FH-RDPG} & $-0.2439$ & $-0.2439$ & $-0.2442$ & $-0.2493$ & $-0.2491$ & $-0.2439$ & $-0.2461$ & $0.0028$\\		
		\hline
	\end{tabular}
\end{table*}  
  
\subsubsection{Performance across $5$ runs}
The individual, average, best observed performance as well as the standard error across $5$ runs are reported in Table \ref{perform}. For each run, the individual performance is obtained by averaging the returns over $100$ test episodes after training is completed, where the returns for MDP and POMDP are calculated according to \eqref{eq13} and \eqref{eq14}, respectively. Note that the individual performance for the myopic, myopic-POMDP, iLQG, and iLQG-POMDP algorithms remain the same across different runs. We can observe that for each run, the individual performance of the proposed FH-DDPG algorithm is always larger (and thus performance is always better) than those of myopic, iLQG, and DDPG algorithms. The maximum performance of FH-DDPG algorithm is larger than those of the myopic, iLQG, and DDPG algorithms by $80\%$, $23\%$ and $18\%$, respectively. Moreover, the average performance of the FH-DDPG algorithm is larger than those of myopic, iLQG, and DDPG algorithms by $80\%$, $22\%$ and $90\%$, respectively. The results demonstrate that the FH-DDPG algorithm performs better than the iLQG algorithm when both algorithms yield approximate solutions to a fully-observable deterministic sequential decision problem with nonlinear dynamics. As shown in Table \ref{perform}, the standard error of the DDPG algorithm is much larger than that of the FH-DDPG algorithm, which indicates that the performance of the proposed FH-DDPG algorithm is much more stable than that of the DDPG algorithm.\par


Similarly, as can be observed from Table \ref{perform}, we can also observe that for each run, the performance of the proposed FH-RDPG algorithm is always larger (and thus performance is always better) than those of myopic-POMDP, iLQG-POMDP, and RDPG algorithms. The maximum performance of FH-RDPG algorithm is larger than those of myopic-POMDP, iLQG-POMDP, and RDPG algorithms by $84\%$, $31\%$, and $26\%$, respectively. Moreover, the average performance of FH-RDPG algorithm is larger than those of myopic-POMDP, iLQG-POMDP, and RDPG algorithms by $84\%$, $30\%$, and $39\%$, respectively. The results demonstrate that the FH-RDPG algorithm performs better than the iLQG-POMDP algorithm when both algorithms yield approximate solutions to a partially-observable deterministic sequential decision problem with nonlinear dynamics. Moreover, FH-RDPG algorithm is more stable than the RDPG algorithm, as the standard error of FH-RDPG is smaller than that of RDPG.\par

  
We can also observe from Table \ref{perform} that the performance of FH-DDPG algorithm are always larger than those of FH-RDPG algorithm. This is due to the fact that FH-DDPG is run in the MDP environment, where the state information is available to the agent for selecting actions; while FH-RDPG algorithm is run in the POMDP environment, which considers the practical scenario that only the observation information is available to the agent. The average performance of FH-RDPG is smaller than that of FH-DDPG by $5\%$. This performance loss of FH-RDPG over FH-DDPG is due to the ``uncertainty for the next time step" as discussed in Remark 3. On the other hand, the average performance of myopic algorithm in POMDP environment is smaller than that in MDP environment by $31\%$. This observation indicates that the FH-RDPG algorithm exploits the history data to make more efficient decisions as compared to the myopic algorithm.\par 

\begin{figure}[!htb]
	\centering
	\includegraphics[width=0.48\textwidth]{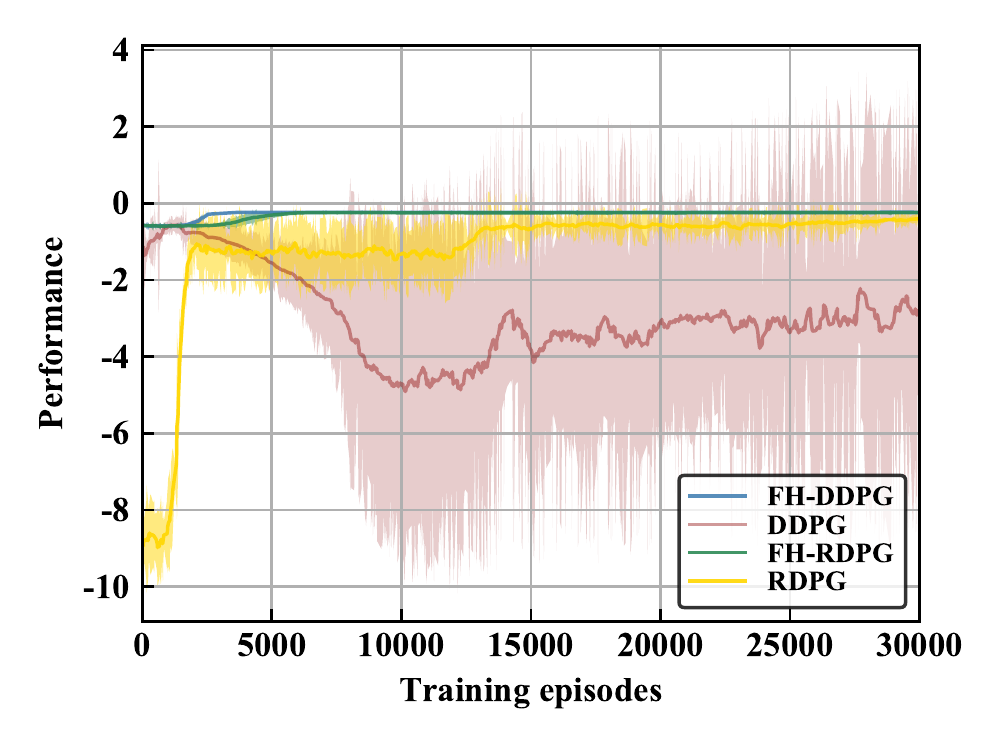}
	\caption{Average performance across 5 runs for the four DRL algorithms. The vertical axis corresponds to the average performance across 5 runs and the shaded areas indicate the standard errors of the four algorithms.}
	\label{convergence}
\end{figure}

\begin{figure*}[!htb]
	\centering
	\subfigure[FH-DDPG]{
		\begin{minipage}[b]{0.2\textwidth}
			\includegraphics[width=1\textwidth]{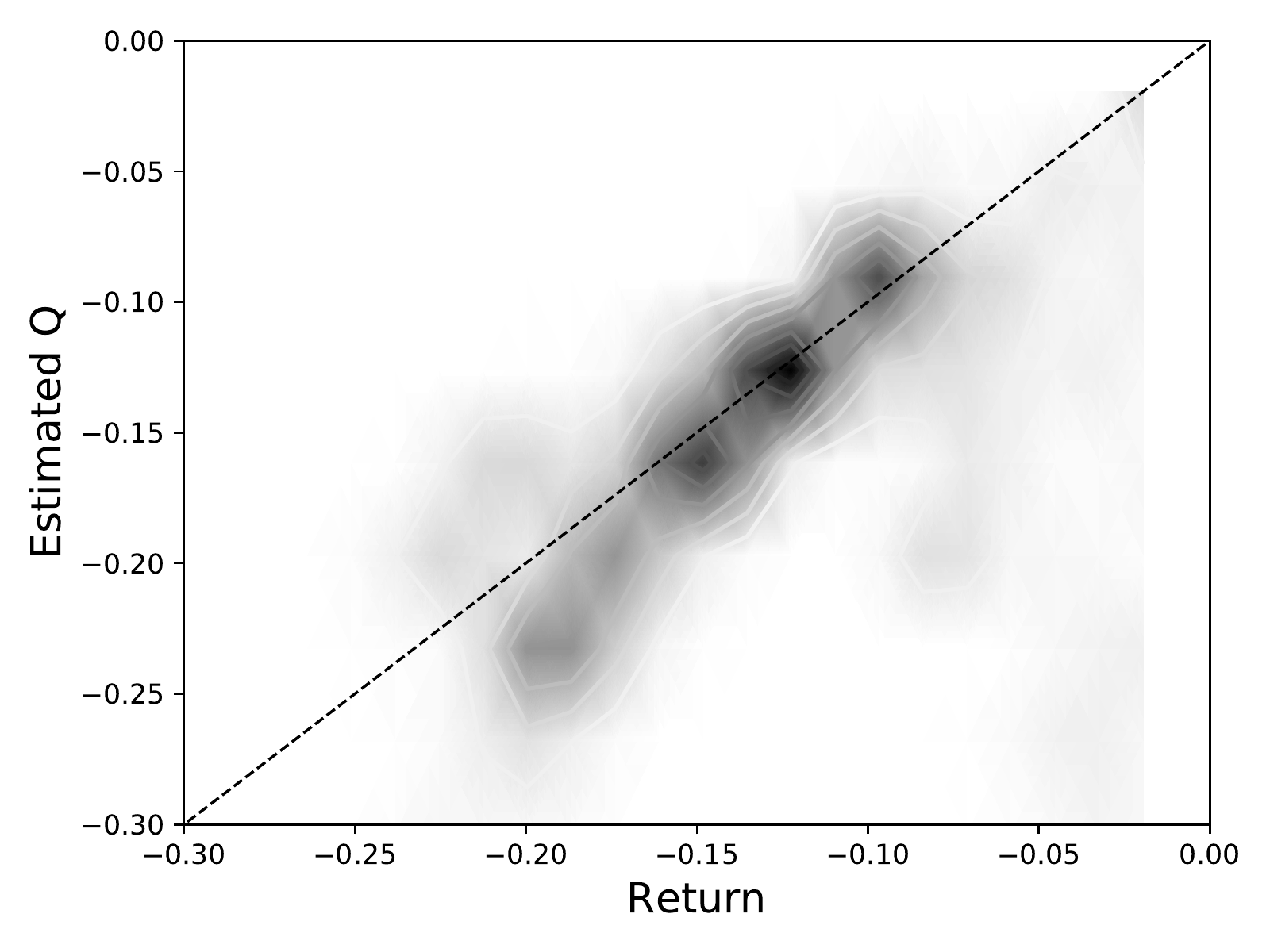}
		\end{minipage}
		\label{q_fhddpg}
	}
	\subfigure[FH-RDPG]{
		\begin{minipage}[b]{0.2\textwidth}
			\includegraphics[width=1\textwidth]{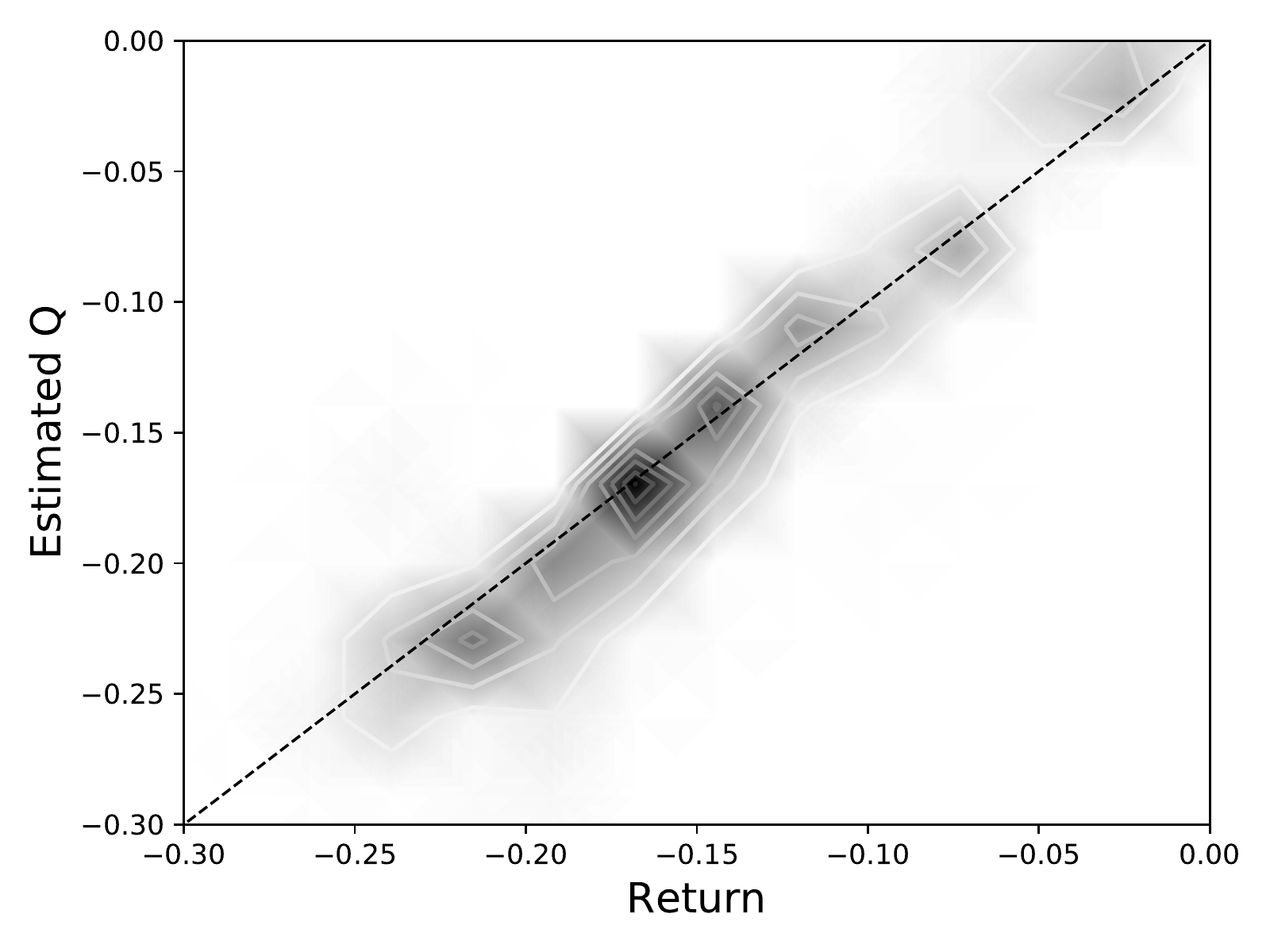}
		\end{minipage}
		\label{q_fhrdpg}
	} 	
	\subfigure[DDPG]{
		\begin{minipage}[b]{0.2\textwidth}
			\includegraphics[width=1\textwidth]{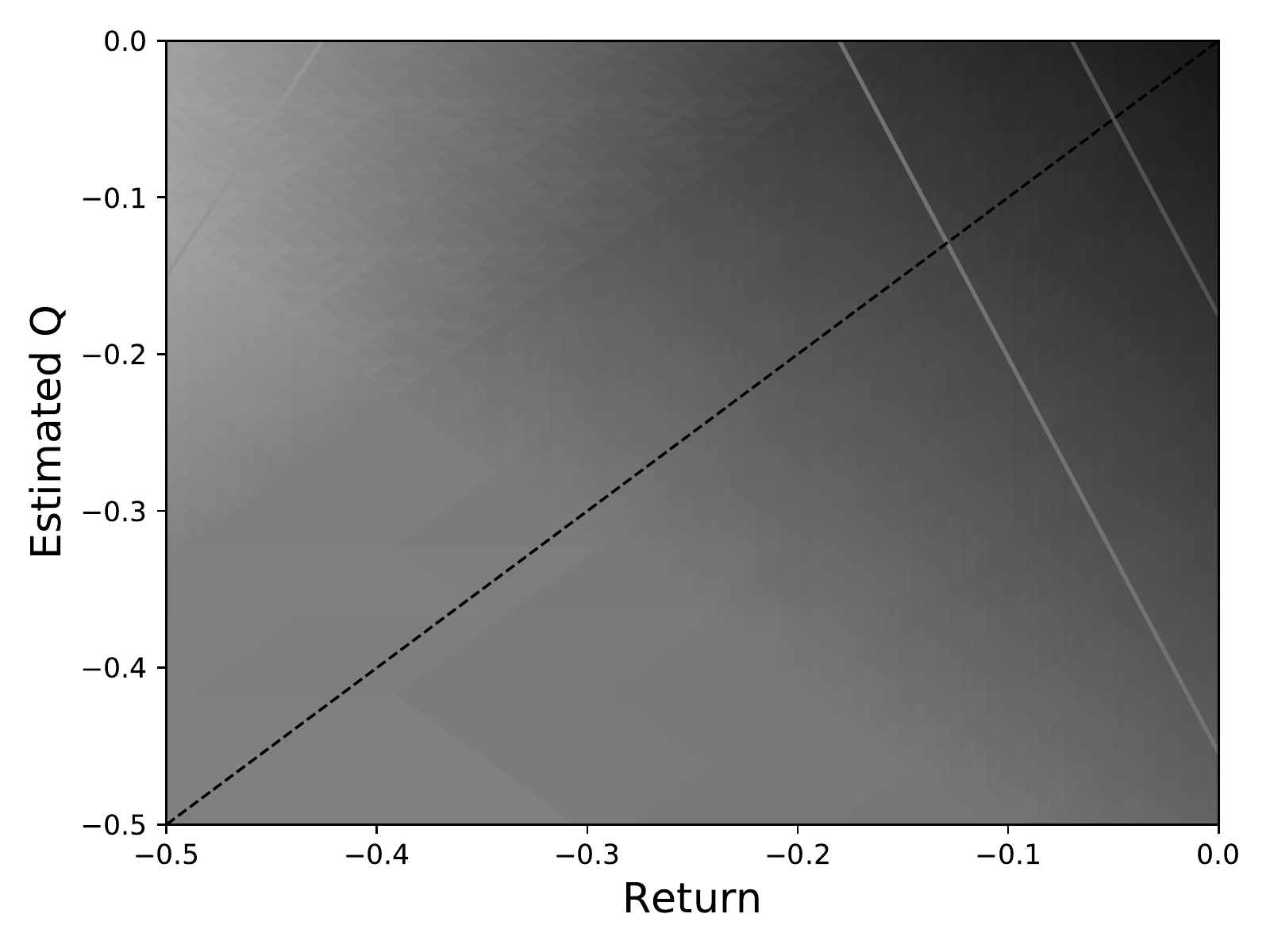}
		\end{minipage}
		\label{q_ddpg}
	} 
	\subfigure[RDPG]{
		\begin{minipage}[b]{0.2\textwidth}
			\includegraphics[width=1\textwidth]{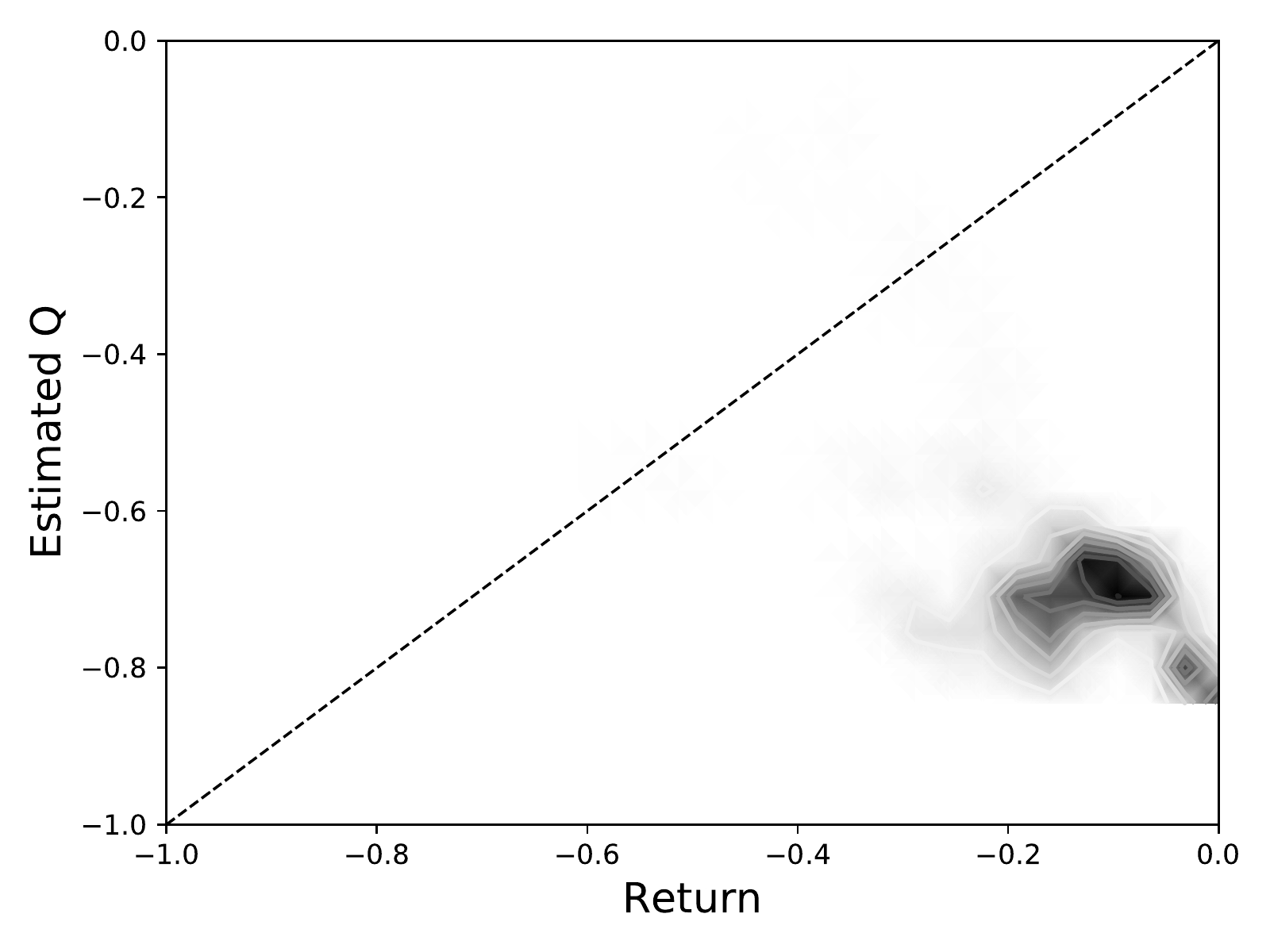}
		\end{minipage}
		\label{q_rdpg}
	} 
	\caption{Density plot showing estimated Q-values versus observed returns from test episodes on 5 runs. The vertical axis corresponds to the estimated Q-values while the horizontal axis corresponds to the true Q-values. }
	\label{q_drl}
\end{figure*}

\begin{figure*}[!t]
	\centering
	\subfigure[FH-DDPG]{
		\begin{minipage}[b]{0.225\textwidth}
			\includegraphics[width=1\textwidth]{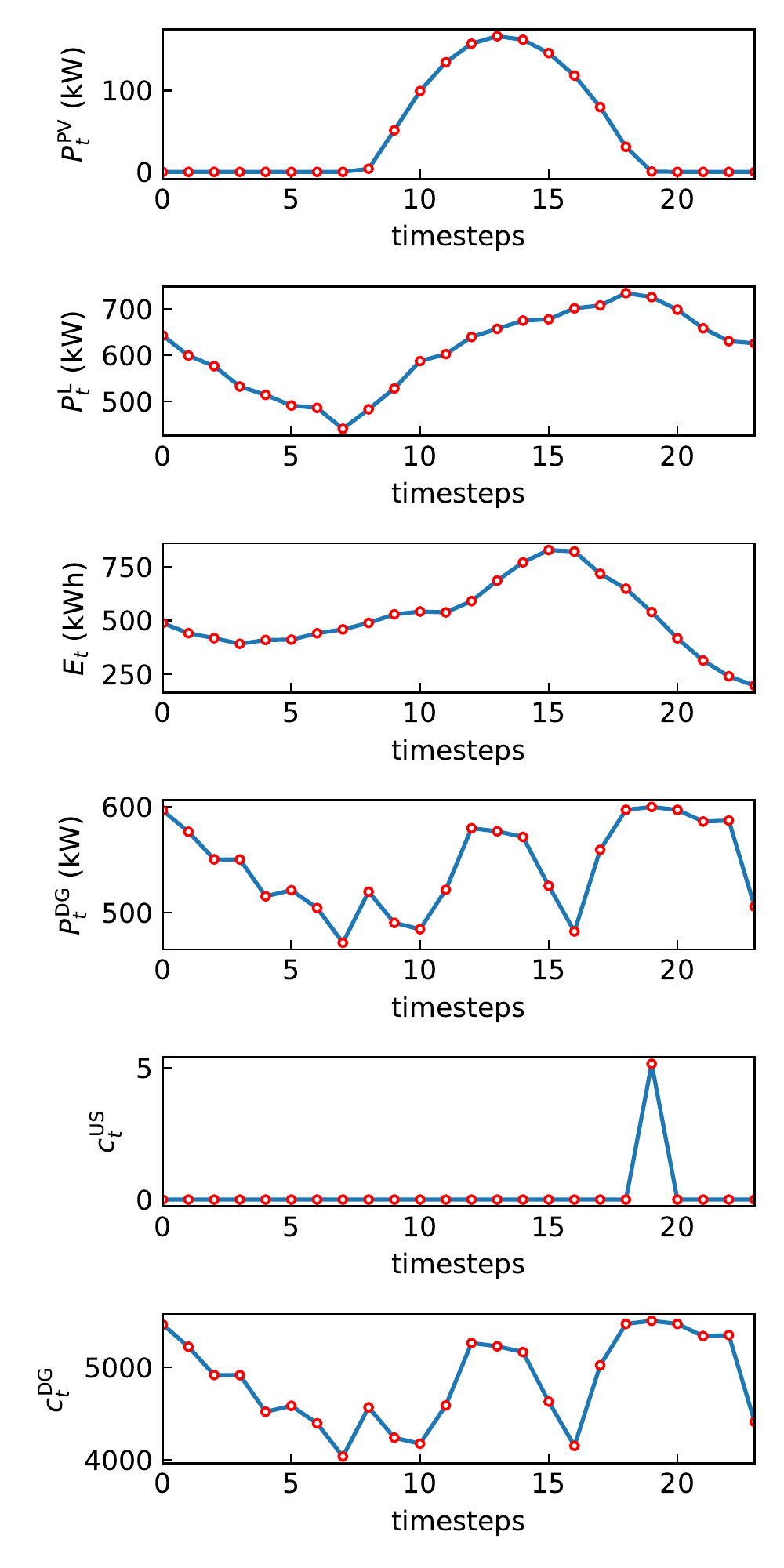}
		\end{minipage}
		\label{fhddpg}
	}	
	\subfigure[DDPG]{
		\begin{minipage}[b]{0.225\textwidth}
			\includegraphics[width=1\textwidth]{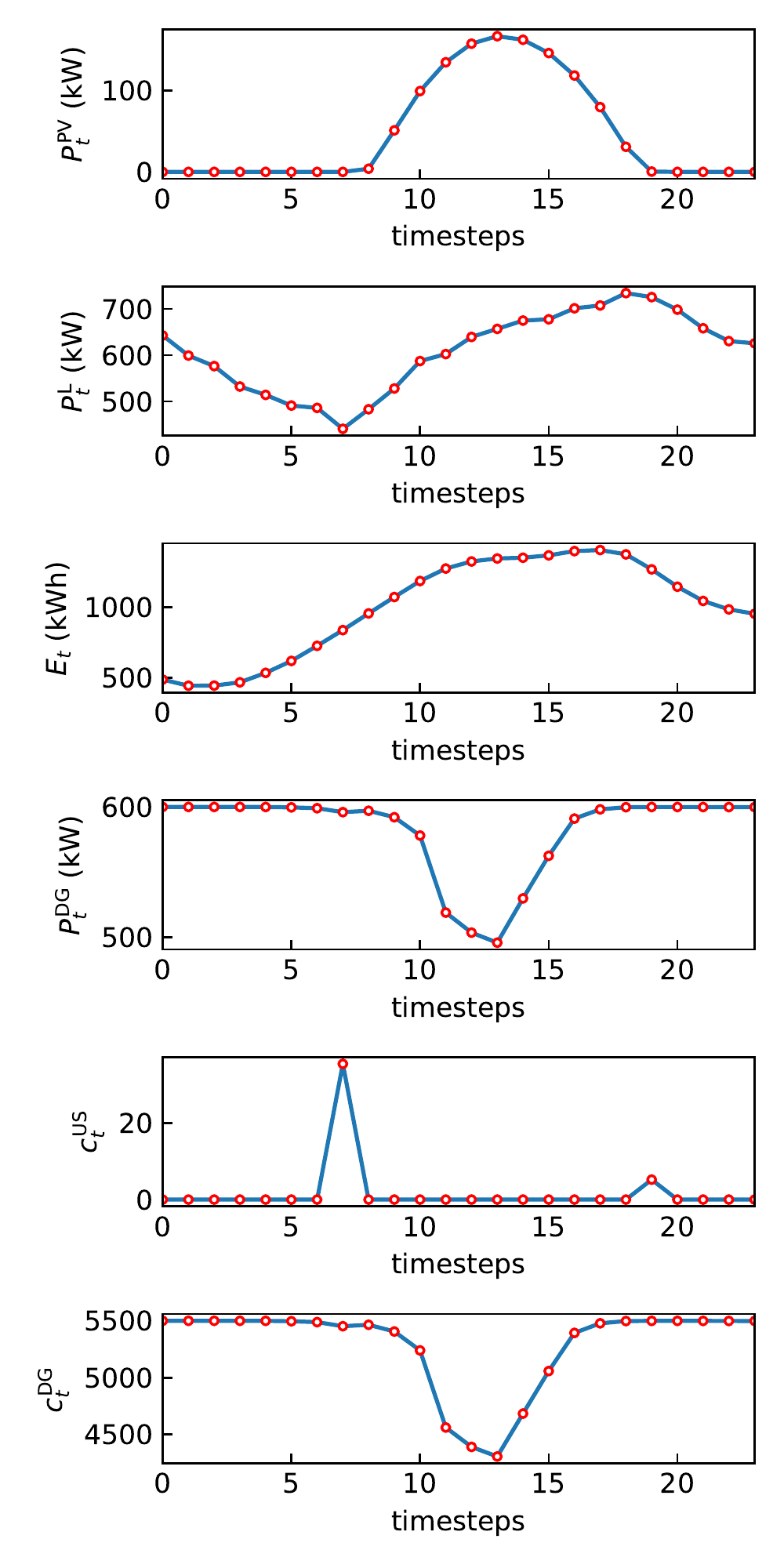}
		\end{minipage}
		\label{ddpg}
	} 
	\subfigure[myopic]{
		\begin{minipage}[b]{0.225\textwidth}
			\includegraphics[width=1\textwidth]{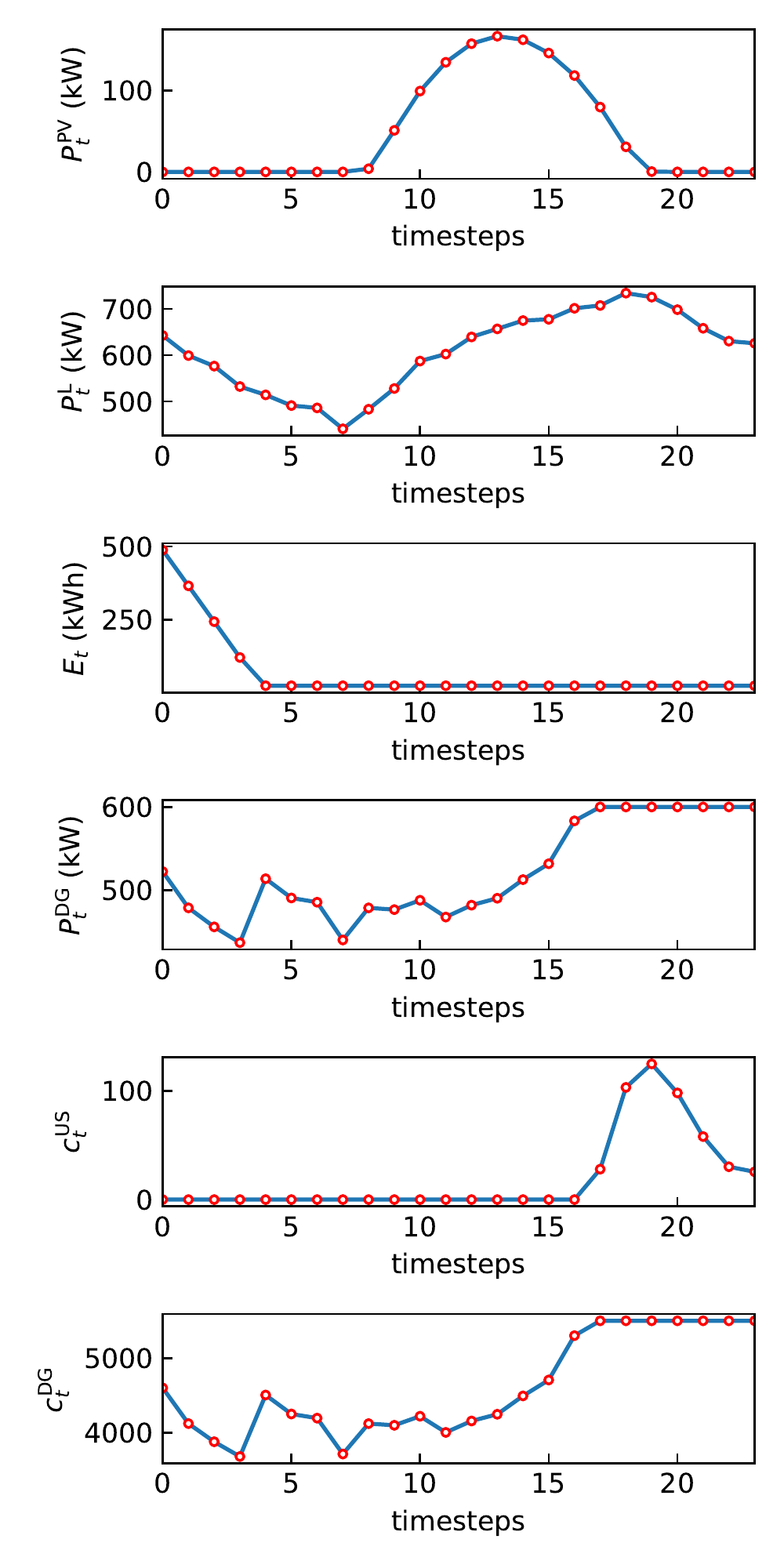}
		\end{minipage}
		\label{myopic}
	}
	\subfigure[iLQG]{
		\begin{minipage}[b]{0.225\textwidth}
			\includegraphics[width=1\textwidth]{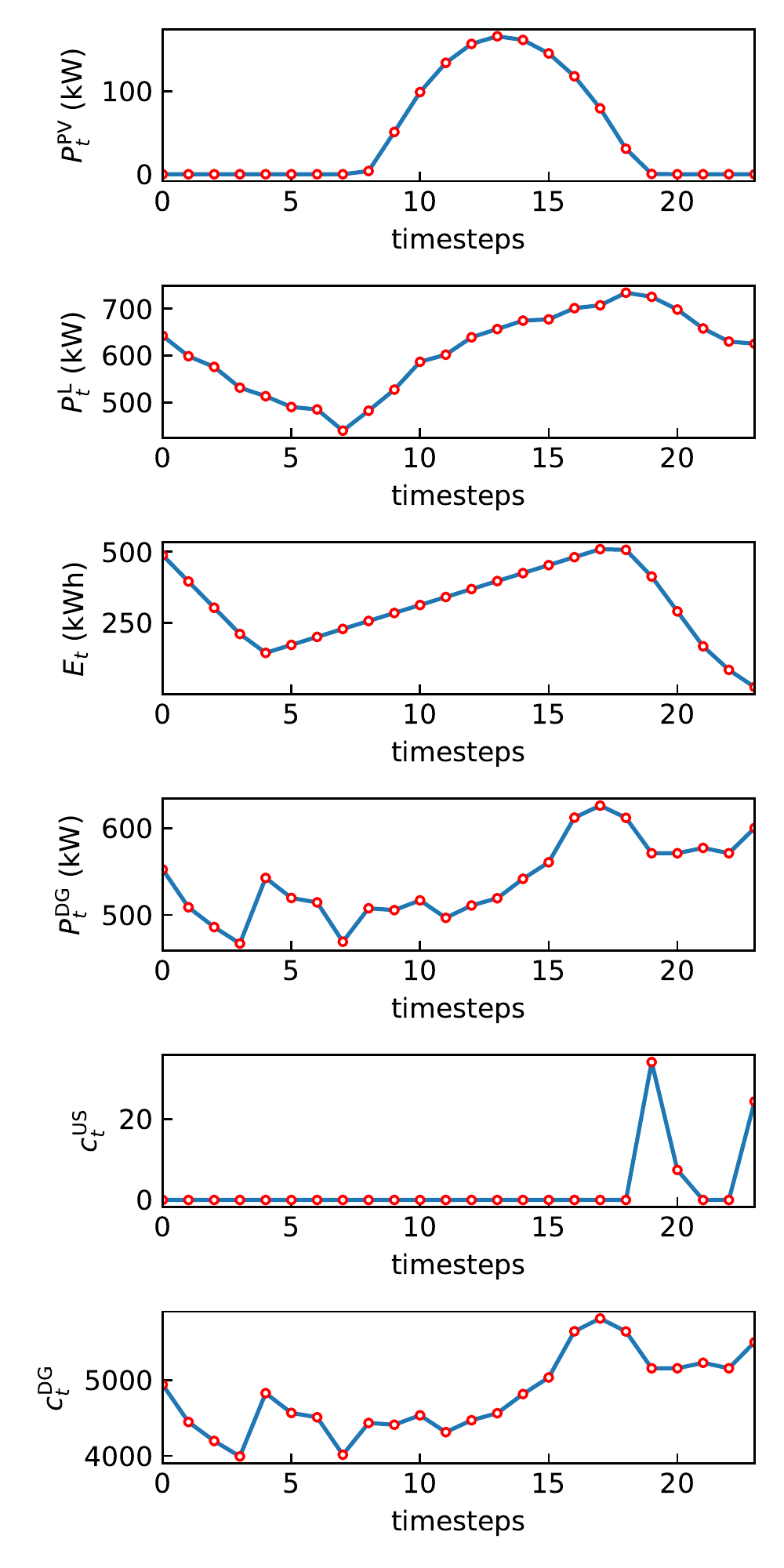}
		\end{minipage}
		\label{ilqg}
	} 
	\caption{Energy dispatch results for a specific test episode under the MDP environment. Each column refers to a one-day energy dispatch result generated by a specific algorithm. In each column, the dynamic energy dispatch results of $P_{t}^{\mathrm{PV}}$, $P_{t}^{\mathrm{L}}$, $E_{t}$, $P_{t}^{\mathrm{DG}}$, $c_{t}^{\mathrm{DG}}$ and $c_{t}^{\mathrm{US}}$ are represented as different curves, respectively.}
	\label{drl_mdp}
\end{figure*}
\subsubsection{Convergence properties}
The policies are evaluated periodically during training without noise. The performance curves for the four DRL algorithms are given in Fig. \ref{convergence}, where the vertical axis corresponds to the average performance across $5$ runs and the shaded areas indicate the standard errors of the four algorithms. For FH-DDPG and FH-RDPG algorithms, we only show the performance curve of the first time step $t=1$ due to space limitation in this paper, as the performance curves for all $T-1$ (for FH-DDPG) or $T$ (for FH-RDPG) time steps are similar, except that the performance grow more and more negative with decreasing time steps. Fig.\ref{convergence} shows that both FH-DDPG and FH-RDPG learn much faster than DDPG and RDPG algorithms. This is due to the fact that FH-DDPG and FH-RDPG algorithms are designed to iteratively train to solve the one-period MDP and POMDP problems with two time steps, which are much easier to converge than the finite-horizon MDP and POMDP problems with $T$ time steps faced by DDPG and RDPG algorithms. Moreover, it can also be observed from Fig.\ref{convergence} that the shaded areas of both FH-DDPG and FH-RDPG algorithms are much smaller than those of the DDPG and RDPG algorithms, indicating that the proposed algorithms perform stably across different runs.  \par

\begin{figure*}[!t]
	\centering
	\subfigure[FH-RDPG]{
		\begin{minipage}[b]{0.225\textwidth}
			\includegraphics[width=1\textwidth]{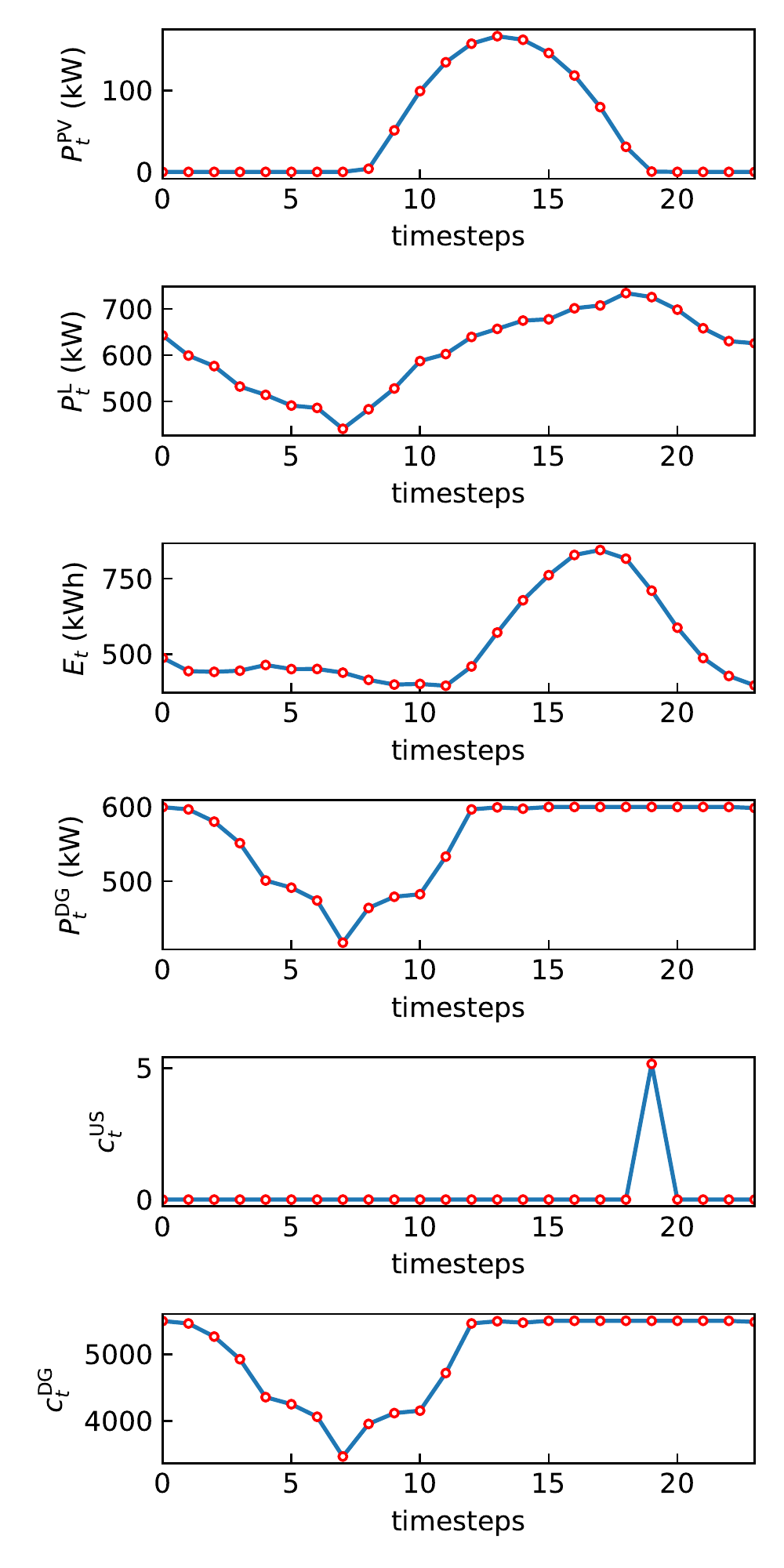}
		\end{minipage}
		\label{fhrdpg}
	}	
	\subfigure[RDPG]{
		\begin{minipage}[b]{0.225\textwidth}
			\includegraphics[width=1\textwidth]{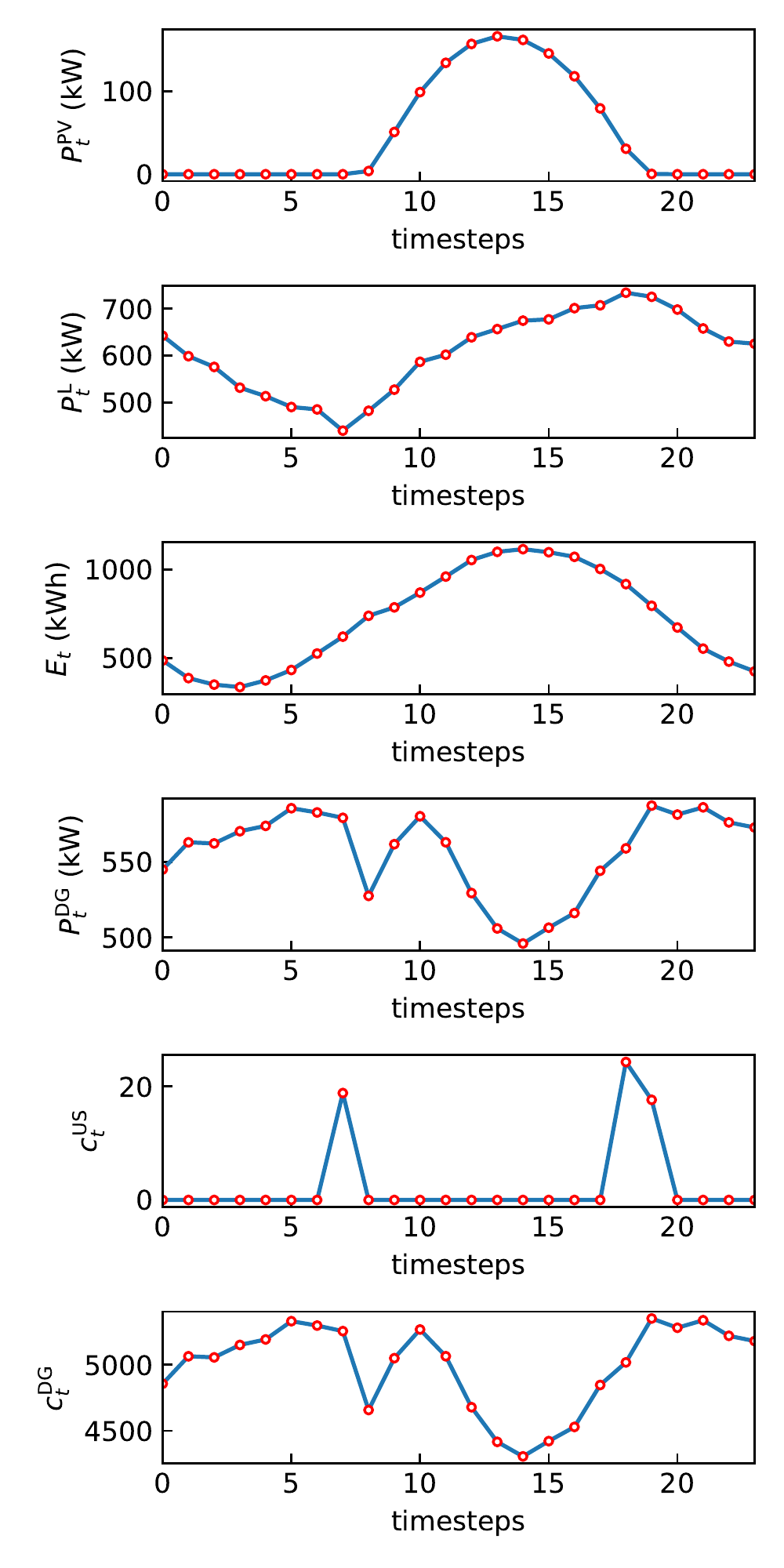}
		\end{minipage}
		\label{rdpg}
	} 
	\subfigure[myopic-POMDP]{
		\begin{minipage}[b]{0.225\textwidth}
			\includegraphics[width=1\textwidth]{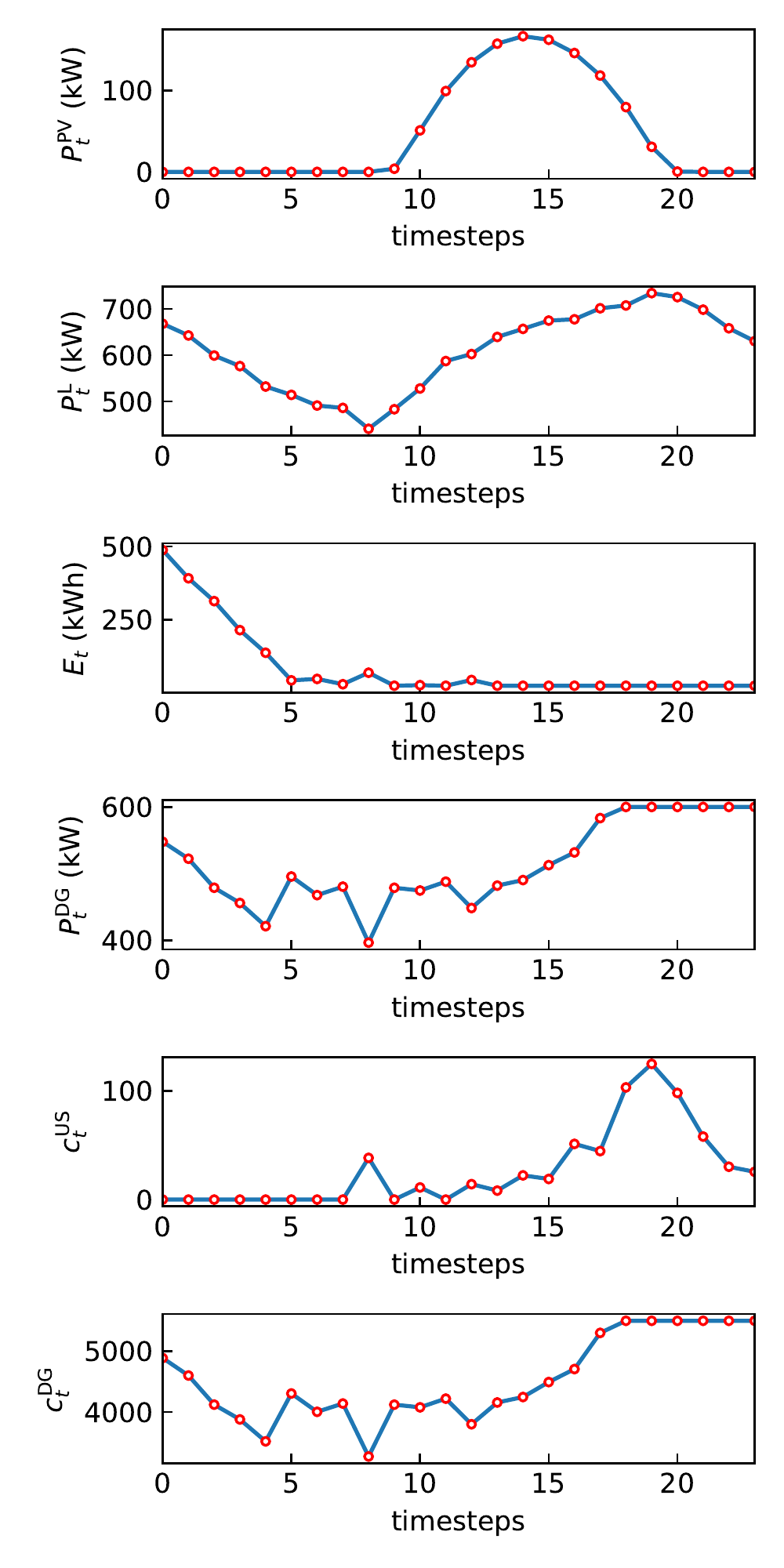}
		\end{minipage}
		\label{myopic_2}
	}
	\subfigure[iLQG-POMDP]{
		\begin{minipage}[b]{0.225\textwidth}
			\includegraphics[width=1\textwidth]{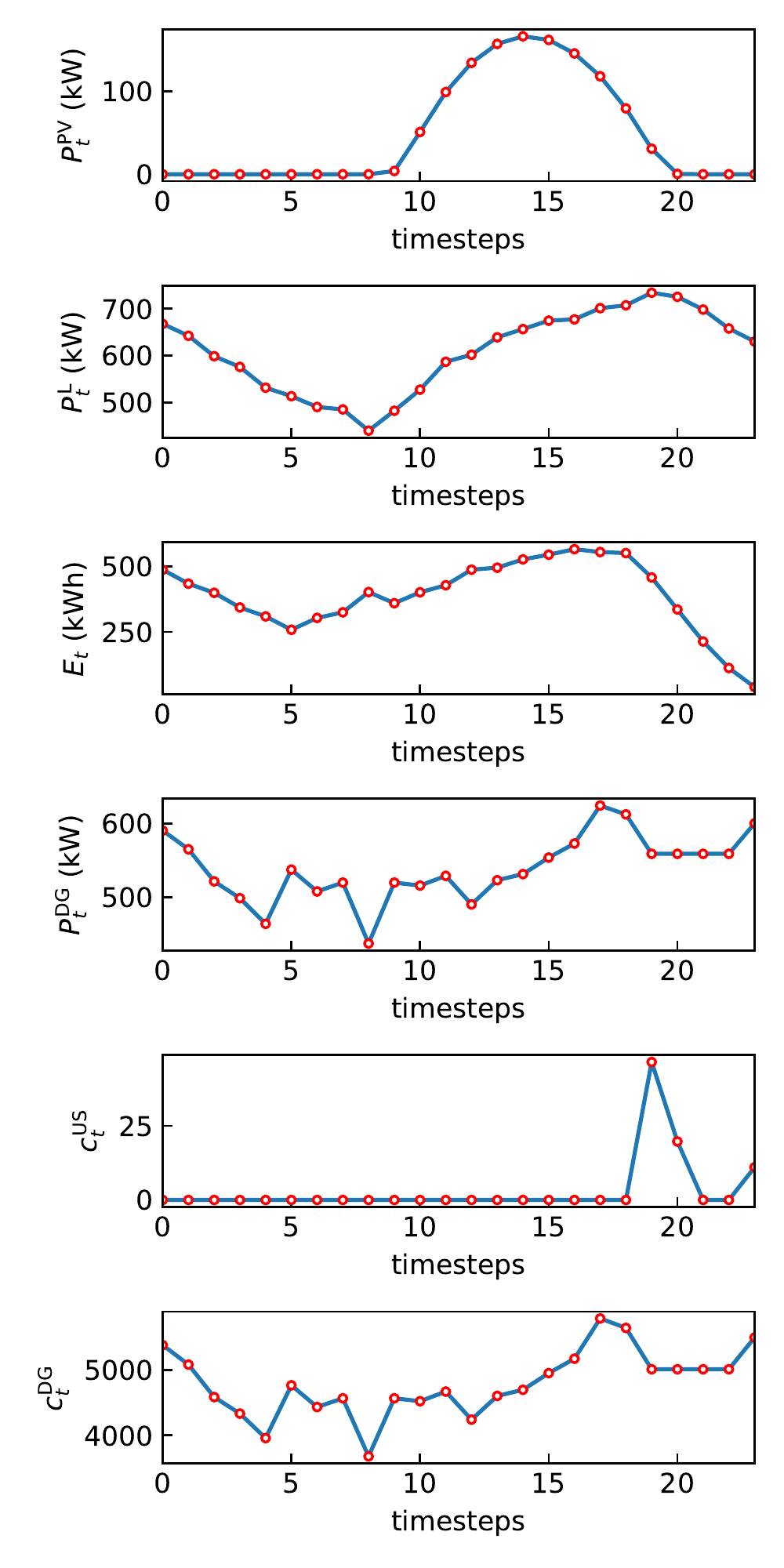}
		\end{minipage}
		\label{ilqg_2}
	}  
	\caption{Energy dispatch results for a specific test episode under the POMDP environment. Each column refers to a one-day energy dispatch result generated by a specific algorithm. In each column, the dynamic energy dispatch results of $P_{t}^{\mathrm{PV}}$, $P_{t}^{\mathrm{L}}$, $E_{t}$, $P_{t}^{\mathrm{DG}}$, $c_{t}^{\mathrm{DG}}$ and $c_{t}^{\mathrm{US}}$ are represented as different curves, respectively.}
	\label{drl_pomdp}
\end{figure*}

\subsubsection{Accuracy of Q-values estimations}
As learning accurate Q-values is very important for the success of actor-critic algorithms, we examined the Q-values estimated by the critic after training for all the four DRL algorithms by comparing them with the true returns seen on the test episodes. Fig. \ref{q_fhddpg} and Fig. \ref{q_fhrdpg} show that both FH-DDPG and FH-RDPG algorithms can provide accurate estimation of Q-values. On the other hand, Fig. \ref{q_rdpg} shows that the estimated Q-values of RDPG have a systematic bias that make the estimated Q-values deviate from the true returns. Finally, Fig. \ref{q_ddpg} demonstrates that the estimated Q-values of DDPG cannot reflect the true returns at all. This is because for the five runs, only Run 1 converges while all the other four runs fail. \par

\subsubsection{Energy dispatch results}
In order to obtain insights for the design of energy dispatch policies, we focus on a specific test episode corresponding to one day, and plot the energy dispatch results $P_{t}^{\mathrm{DG}}$ along with the immediate DG generation cost $c_{t}^{\mathrm{DG}}$ and power unbalance cost $c_{t}^{\mathrm{US}}$ as well as PV and load data for all the time steps $t\in\{1,2,\cdots,24\}$. Fig. \ref{drl_mdp} shows the results for FH-DDPG, DDPG, myopic and iLQG algorithms under the MDP environment. The day starts at around $12:00$ AM midnight, when the PV generation is zero. The PV output power starts to increase from $8:00$ AM until it reaches the peak value at around $2:00$ PM. The load decreases from midnight till around $7:00$ AM in the morning, and starts to increase until the peak value is reached at $6:00$ PM. The initial SoC of battery is $500$kWh.\par

As we can see in Fig. \ref{fhddpg}, in the energy dispatch policy obtained by the FH-DDPG algorithm, the DG output power first tracks the load demand so that the SoC of the battery remains almost constant. After the PV generation starts to increase after $8:00$ AM, the DG power output slowly charges the battery so that the SoC increases to $750$kWh. This is very important, as when the load demand reaches the peak value at $6:00$ PM later that day, the PV power generation is no longer available, and the maximum DG output power is not enough to meet the large load demand. In this case, the battery can discharge to maintain the power supply and demand balance. As a result, the cost of power unbalance $c_{t}^{\mathrm{US}}$ is always $0$ except for time step $19$. Moreover, note that little DG output power is wasted to charge up the battery, as the SoC of the battery at the end of the day is at a low level. Therefore, DG generation cost $c_{t}^{\mathrm{DG}}$ is also low. \par 

On the other hand, as can be observed from Fig. \ref{myopic}, the myopic algorithm quickly discharges the battery energy at the beginning of the day in order to save DG output power, without considering that the DG output power will not be able to meet load demand later that day. This results in that the cost of power unbalance $c_{t}^{\mathrm{US}}$ is not $0$ after $4:00$ PM. In contrast, DDPG algorithm begins to charge up the battery at the beginning of the day as shown in Fig. \ref{ddpg}. Although the battery can help to maintain power balance in this case, the DG output power is wasted as the SoC of battery is quite high at the end of the day, resulting in large DG power generation cost $c_{t}^{\mathrm{DG}}$. As shown in Fig. \ref{ilqg}, the iLQG algorithm discharges the battery energy to meet the load demand for a few hours at the beginning of the episode. Then, at $4:00$ AM it starts to charge the battery to prevent the upcoming energy shortage. The SoC falls to the minimum level at the end of the episode, which efficiently reduces waste of energy. However, there are a few time steps with power unbalance at the end of the day.\par

Fig. \ref{drl_pomdp} shows the results for FH-RDPG, RDPG, myopic-POMDP and iLQG-POMDP algorithms under the POMDP environment. As can be observed in Fig. \ref{fhrdpg} and Fig. \ref{rdpg}, both FH-RDPG and RDPG, algorithms charge the battery before the peak hour for the loads similar to FH-DDPG and DDPG algorithms, so that they can avoid power unbalance in most time steps. On the other hand, the myopic algorithm under POMDP environment discharges the battery from the beginning of the day similar to its behaviour under MDP environment, which results in power unbalance later that day. The iLQG-POMDP algorithm acts similarly with iLQG algorithm and achieves relatively small cost. FH-RDPG algorithm keeps the power balanced, i.e., $c_{t}^{\mathrm{US}}=0$, in more time steps than RDPG algorithm. On the other hand, the SoC of battery at the end of day for both FH-RDPG and RDPG algorithms are larger than that of FH-DDPG algorithm, resulting in larger DG power generation cost $c_{t}^{\mathrm{DG}}$.\par

\subsubsection{Impact of coefficients $k_{1}$ and $k_{2}$}
\begin{figure}[!htb]
	\centering
	\includegraphics[width=0.42\textwidth]{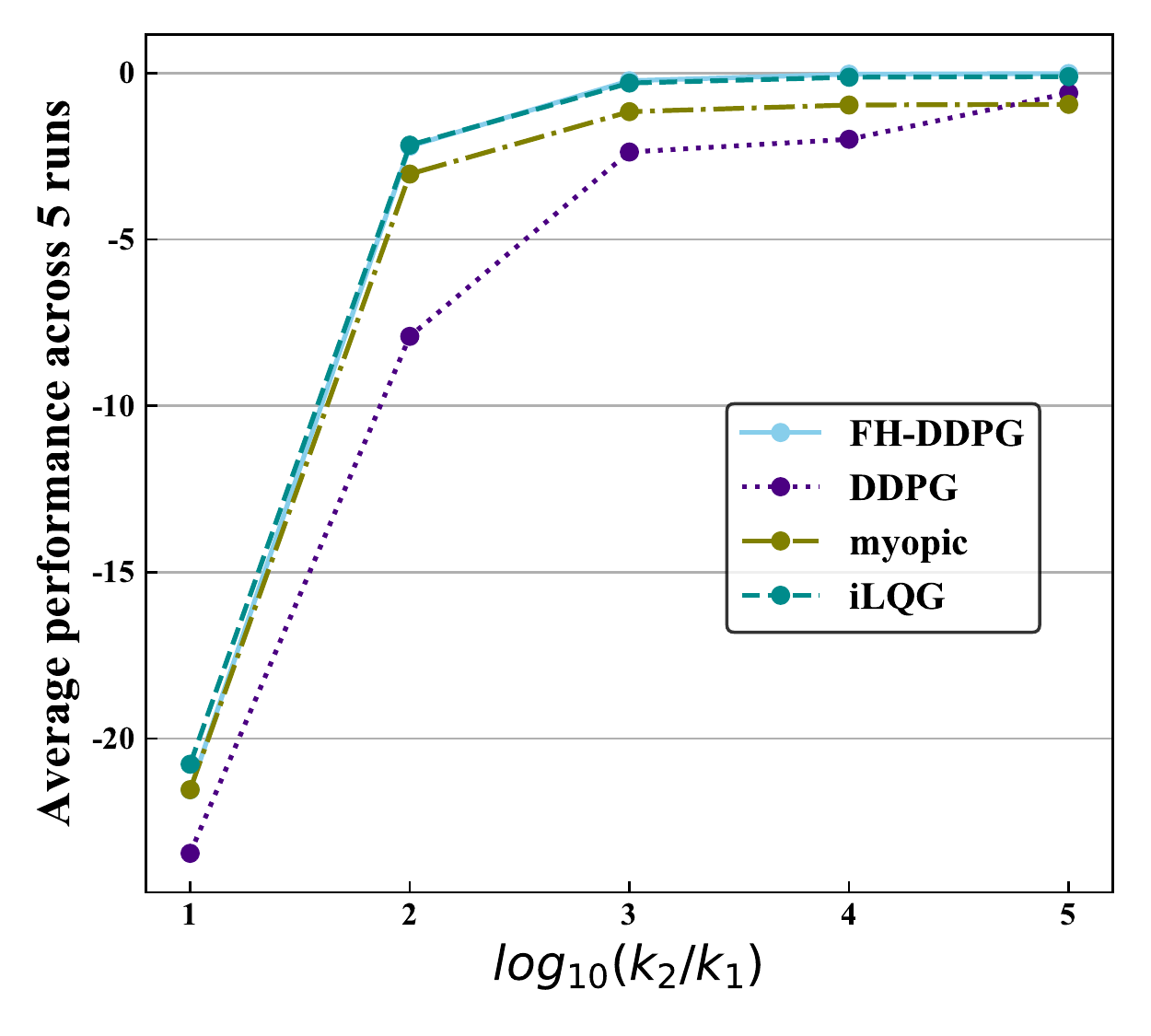}
	\caption{Performance obtained by averaging the returns over $100$ episodes for different ratios of $k_{2}$ to $k_{1}$. The ratios are set to be $10$, $10^2$, $10^3$, $10^4$, and $10^5$ respectively, which correspond to different weights of the power generation cost and power unbalance cost.}
	\label{k1k2_total}
\end{figure}

\begin{figure}[!htb]
	\centering
	\includegraphics[width=0.48\textwidth]{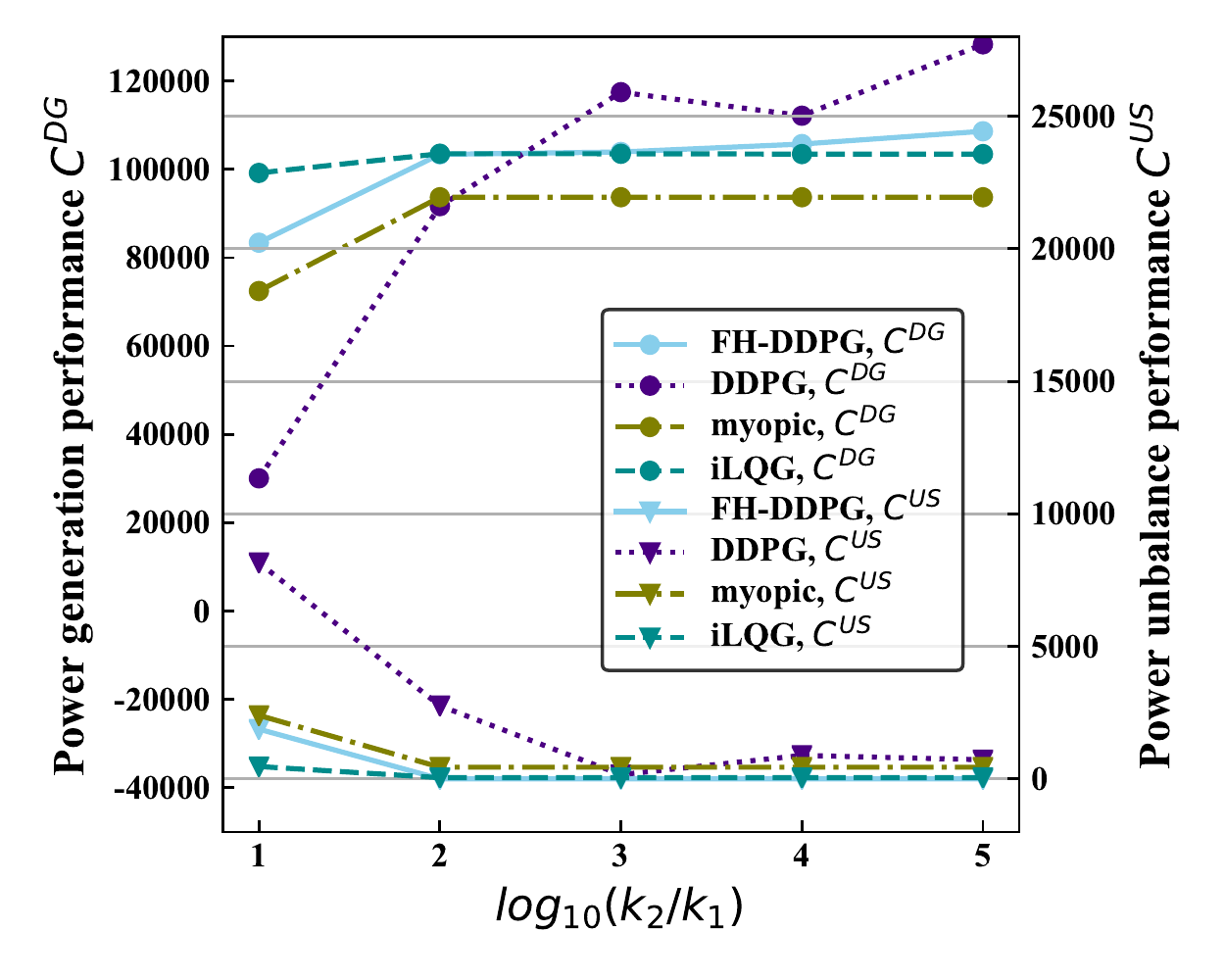}
	\caption{Power generation performance $C^{DG}$ and power unbalance performance $C^{US}$, which are obtained by averaging the sum of power generation costs in one day and the sum of power unbalance costs in one day over 100 test episodes. The ratios of $k_{2}$ to $k_{1}$ are set to be $10$, $10^2$, $10^3$, $10^4$, and $10^5$ respectively, which correspond to different weights of the power generation cost and power unbalance cost.}
	\label{k1k2_cdg_cus}
\end{figure}

The reward function as defined in \eqref{eq10} is the sum of two components, i.e., the cost of power generation $c_{t}^{\mathrm{DG}_{d}}$ and the cost of power unbalance $c_{t}^{\mathrm{US}}$. As optimizing these two components can be conflicting, we use coefficients $k_{1}$ and $k_{2}$ to indicate the relative importance of these two components. In order to evaluate the impact of $k_{1}$ and $k_{2}$ on performance, we set the ratio of $k_{2}$ to $k_{1}$ to be $10$, $10^2$, $10^3$, $10^4$, and $10^5$, respectively, and the performance of the FH-DDPG, DDPG, myopic, and iLQG algorithms are shown in Fig. \ref{k1k2_total}. Moreover, the power generation performance $C^{\mathrm{DG}}$ and power unbalance performance $C^{\mathrm{US}}$ are shown in Fig. \ref{k1k2_cdg_cus}, which are obtained by averaging the sum of power generation costs in one day, i.e., $ \sum_{t=1}^{T}\sum_{d=1}^{D}c_{t}^{\mathrm{DG}_{d}}$ and the sum of power unbalance costs in one day, i.e., $\sum_{t=1}^{T}c_{t}^{\mathrm{US}}$, over $100$ test episodes. It can be observed from Fig. \ref{k1k2_total} that FH-DDPG algorithm achieves the best performance over different ratios of $k_{2}$ to $k_{1}$. From Fig. \ref{k1k2_cdg_cus}, it can be observed that with increasing ratios of $k_{2}$ to $k_{1}$, the power unbalance cost $C^{\mathrm{US}}$ becomes smaller while the power generation cost $C^{\mathrm{DG}}$ becomes larger, as minimizing $C^{\mathrm{US}}$ is given higher priority over minimizing $C^{\mathrm{DG}}$ in the design of the reward function. However, notice that when the ratios of $k_{2}$ to $k_{1}$ reaches $10^3$, further increasing the ratio has insignificant impact on the performance, as the weighted power unbalance cost is already much larger than the weighted power generate cost in the reward. \par

\subsection{Train with history data (Case III and Case IV)}
In order to achieve a good performance, the trained policy based on past data needs to be general enough to adapt to the unknown situation in the future. Therefore, we need to train the DRL algorithms based on the past data in multiple days to prevent the learned policy from over-fitting to the statistics of one particular day. \par

\begin{figure}[!htb]
	\centering
	\includegraphics[width=0.48\textwidth]{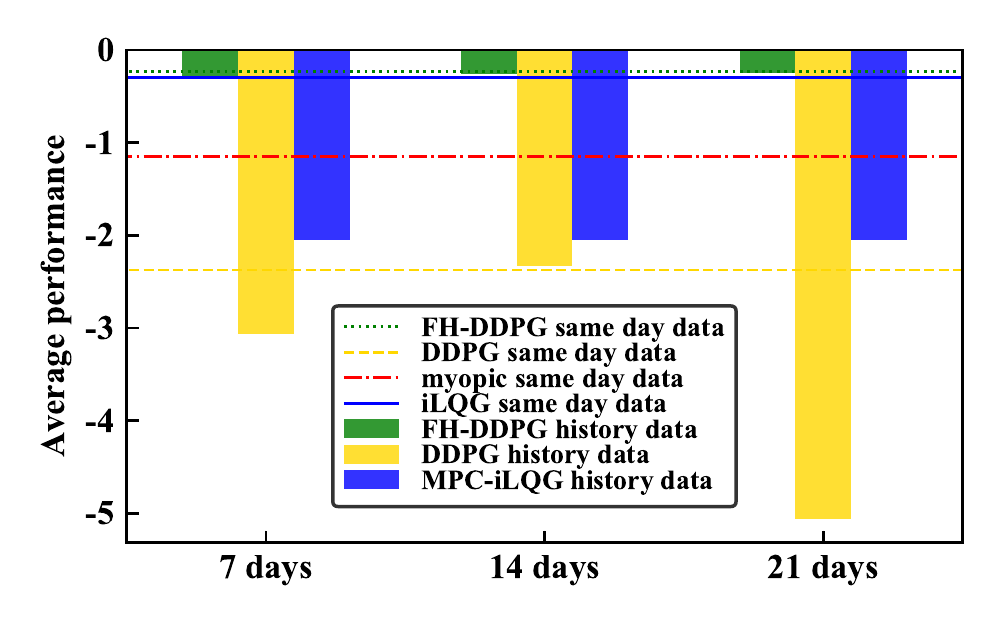}
	\caption{Average performance of FH-DDPG and DDPG when trained based on history data and same-day data across $5$ runs. The green bar on the left represents the FH-DDPG algorithm, while the yellow bar in the middle represents the DDPG algorithm. The blue bar on the right represents the MPC-iLQG algorithm, where the PV and load are predicted based on the past $100$ days' data. As for the horizontal lines, the green, yellow, red and blue line refers to the average performance obtained by FH-DDPG, DDPG, myopic and iLQG algorithm across $5$ runs, respectively, after training on the same-day data.}
	\label{mdp}
\end{figure}

\begin{figure}[!htb]
	\centering
	\includegraphics[width=0.48\textwidth]{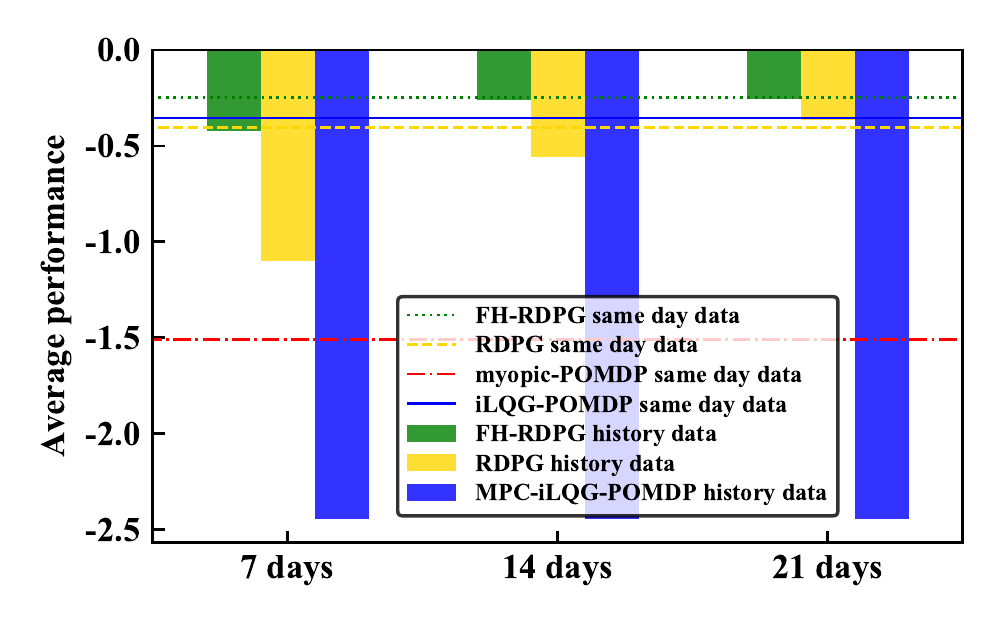}
	\caption{Average performance of FH-RDPG and RDPG across $5$ runs when trained based on history data and same-day data. The green bar on the left represents the FH-RDPG algorithm, while the yellow bar in the middle represents the RDPG algorithm. The blue bar on the right represents the MPC-iLQG-POMDP algorithm, where the PV and load are predicted based on the past $100$ days' data. The green, yellow, red and blue line refers to the average performance across $5$ runs obtained by FH-RDPG, RDPG, myopic-POMDP and iLQG-POMDP algorithm, respectively, after training on the same-day data.}
	\label{pomdp}
\end{figure}

We report the average performance of the four DRL algorithms across $5$ runs where the policies are trained based on the load and PV data of past $7$, $14$, and $21$ days, respectively. Specifically, the PV and load data from 23:59 PM on June 30, 2017 to 22:59 PM on July 07, 2017 (7 days), from 23:59 PM on June 23, 2017 to 22:59 PM on July 07, 2017  (14 days), and from 23:59 PM on June 16, 2017 to 22:59 PM on July 07, 2017  (21 days) are used to train the algorithms, while the data from 23:59 PM on July 07, 2017 to 22:59 PM on July 08, 2017 are used to test the algorithms. We also report the results of MPC-iLQG and MPC-iLQG-POMDP algorithms as explained in Section V.A. The prediction for the future PV and load data is based on the data from the past $100$ days using RNN, and the model prediction errors for both MPC algorithms are represented by mean square error (MSE) after normalization, which are $1.39\times10^{-3}$ for $P_{t}^{\mathrm{L}}$ and $2.27\times10^{-3}$ for $P_{t}^{\mathrm{PV}}$, respectively. \par

Fig. \ref{mdp} and Fig. \ref{pomdp} show the results when the trained policies are tested on the current day in MDP and POMDP environments, respectively. It can be observed that the proposed FH-DDPG and FH-RDPG algorithms (green bars) still outperform the baseline DDPG and RDPG algorithms (yellow bars) by a large margin when trained based on history data. Moreover, the number of days to train the DRL algorithms has a larger impact on DDPG and RDPG algorithms than on FH-DDPG and FH-RDPG algorithms. FH-DDPG and FH-RDPG algorithms also achieve better performance compared with MPC-iLQG and MPC-iLQG-POMDP algorithms (blue bars), respectively. \par

We also compare the above results with those in Table \ref{perform}, where the policies are trained based on the same-day data instead of history data. For comparison purpose, we plot the results in Table \ref{perform} by four lines with different styles in Fig. \ref{mdp} and Fig. \ref{pomdp}. It can be observed that the performance of FH-DDPG and FH-RDPG are quite similar when trained based on history data and same-day data. This is also true for DDPG and RDPG algorithms. This result shows that the DRL algorithms can cope well with the ``uncertainty for the next day" as discussed in Remark 3. The reason is that when the DRL algorithms are trained using history data, sub-optimal policies are learned for a stochastic sequential decision problem where the dynamics of the system, in particular evolution of the PV and load is considered as a stochastic model. The DRL algorithms work well as long as the environment is stationary where the stochastic properties of the history (training) data and test data are similar. Therefore, while a good algorithm is generally not expected to generate close training error and testing error in supervised learning using deep neural networks, this is not the case for DRL, as DRL uses the history data to simulate a stochastic environment instead of learning a function approximation. On the other hand, the performance gap between iLQG (resp. iLQG-POMDP) algorithm and its MPC counterpart MPC-iLQG (resp. MPC-iLQG-POMDP) is relatively large. If prediction errors of PV and load can be reduced by better prediction models, it is expected that better performance can be achieved for MPC-iLQG and MPC-iLQG-POMDP algorithms. This improvement deserves more exploration, but it is out of the scope of this paper. \par

%
%
    
\section{Conclusion}
This paper studied DRL-based dynamic energy dispatch methods for IoT-driven smart isolated MG. Firstly, a finite-horizon MDP model and a finite-horizon POMDP model were formulated for energy dispatch, respectively. Then, considering the characteristics of finite-horizon setting, FH-DDPG and FH-RDPG algorithms were proposed to provide better stability performance over the baseline DRL algorithms under MDP and POMDP models, respectively. Finally, the performance of FH-DDPG and FH-RDPG algorithms were compared with DDPG, RDPG, iLQG, and myopic algorithms based on real same-day data and history data in MG. The results show that our proposed algorithms have derived better energy dispatch policies for both MDP and POMDP models. Moreover, by comparing between FH-DDPG and FH-RDPG algorithms, as well as the results of training with same-day data and history data, we quantified the impact of uncertainty on system performance. We have also demonstrated that the FH-RDPG algorithm can make efficient decisions when only partial state information is available as in most practical scenarios. Our future work involves the joint optimization of unit commitment and energy dispatch of multiple DGs in isolated MGs. Moreover, we will explore the zero/one/few-shot studies to further improve the generalization capacity of the DRL algorithms when trained with history data. \par


\bibliography{microgridDRL}{}
\bibliographystyle{IEEEtran}

\end{document}